%% file: main.tex
\documentclass[10pt,twocolumn,letterpaper]{article}

\usepackage{wacv}              


\definecolor{wacvblue}{rgb}{0.21,0.49,0.74}
\usepackage{multirow}
\usepackage{wrapfig}
\usepackage{adjustbox}
\usepackage{pifont}
\usepackage{tabularx}
\usepackage[table]{xcolor}
\usepackage{makecell}
\definecolor{lightblue}{rgb}{0.84,0.91,0.97}
\newcommand{\cmark}{\ding{51}}%
\newcommand{\xmark}{\ding{55}}%
\newcommand{\q}[1]{`#1'}
\newcommand{\dq}[1]{``#1''}

\newcommand{\jh}[1]{{\color{black}{#1}}}
\newcommand{\rev}[1]{{\color{black}{#1}}}
\usepackage[pagebackref,breaklinks,colorlinks,allcolors=wacvblue]{hyperref}

\def\wacvPaperID{779} 
\def\confName{WACV}
\def\confYear{2026}

\title{Relevance-aware Multi-context Contrastive Decoding \\ for Retrieval-augmented Visual Question Answering}

\author{Jongha Kim\\
Korea University\\
{\tt\small jonghakim@korea.ac.kr}
\and
Byungoh Ko\\
Korea University\\
{\tt\small ko990128@korea.ac.kr}
\and
Jeehye Na\thanks{This work was conducted at Korea University.}\\
KAIST\\
{\tt\small jeehyena@kaist.ac.kr}
\and
Jinsung Yoon\\
Google Cloud AI\\
{\tt\small jinsungyoon@google.com}
\and
Hyunwoo J. Kim\thanks{Corresponding author.}\\
KAIST\\
{\tt\small hyunwoojkim@kaist.ac.kr}
}

\begin{document}
\maketitle
\input{sec/0_abstract}    
\input{sec/1_intro}
\input{sec/2_related_works}
\input{sec/3_method}
\input{sec/4_experiments}
\input{sec/5_analysis}
\input{sec/6_conclusion}

\section*{Acknowledgements}
This work was partly supported by Institute for Information \& Communications Technology Promotion (IITP) grant funded by the Korea government (MSIP) (No. RS-2024-00443251, Accurate and Safe Multimodal, Multilingual Personalized AI Tutors, 40\%), National Research Foundation of Korea (NRF) grant funded by the Korea government (MSIT) (NRF-2023R1A2C2005373, 30\%), and Virtual Engineering Platform Project (Grant No. P0022336, 30\%), funded by the Ministry of Trade, Industry \& Energy (MoTIE, South Korea).

{
    \small
    \bibliographystyle{ieeenat_fullname}
    \bibliography{main}
}

\input{sec/X_suppl}

\end{document}

%% file: sec/0_abstract.tex
\begin{abstract}
Despite the remarkable capabilities of Large Vision Language Models (LVLMs), they still lack detailed knowledge about specific entities.
Retrieval-augmented Generation (RAG) is a widely adopted solution that enhances LVLMs by providing additional contexts from an external Knowledge Base.
However, we observe that previous decoding methods for RAG are sub-optimal as they fail to sufficiently leverage multiple relevant contexts and suppress the negative effects of irrelevant contexts.
To this end, we propose \textbf{R}elevance-aware \textbf{M}ulti-context \textbf{C}ontrastive \textbf{D}ecoding (RMCD), a novel decoding method for RAG.
RMCD outputs a final prediction by combining outputs predicted with each context, where each output is weighted based on its relevance to the question.
By doing so, RMCD effectively aggregates useful information from multiple relevant contexts while also counteracting the negative effects of irrelevant ones.
Experiments show that RMCD consistently outperforms other decoding methods across multiple LVLMs, achieving the best performance on three knowledge-intensive visual question-answering benchmarks.
Also, RMCD can be simply applied by replacing the decoding method of LVLMs without additional training.
Analyses also show that RMCD is robust to the retrieval results, consistently performing the best across the weakest to the strongest retrieval results.
Code is available at \url{https://github.com/mlvlab/RMCD}.
\end{abstract}

%% file: sec/1_intro.tex
\section{Introduction}
Despite the remarkable \jh{performance} of Large Vision Language Models (LVLMs)~\cite{li2023blip, dai2024instructblip, liu2024visual, liu2023improved} in various vision-language tasks, including image captioning~\cite{lin2014microsoft,agrawal2019nocaps}, visual question answering~\cite{goyal2017making,marino2019ok,hudson2019gqa} \jh{and} more, these models often lack detailed knowledge about specific objects. 
Pointing out the problem, knowledge-intensive visual question answering (VQA) benchmarks~\cite{marino2019ok,shah2019kvqa,jain2021select,chen2023can,mensink2023encyclopedic,hu2023open} have been proposed, requiring detailed knowledge about specific images. 
For example, a question might be, \textit{\dq{When was this tower built?}} when the Eiffel Tower is present in an image.

\input{tab_fig_tex/motivatoin_fig}
Retrieval-augmented Generation (RAG)~\cite{lewis2020retrieval} is a widely adopted solution to the problem, enhancing \jh{LVLMs} by providing additional contexts retrieved from an external Knowledge Base (KB). 
A typical RAG pipeline consists of two stages: a retrieval stage and a generation stage.
\jh{In the retrieval stage, multiple contexts with the highest retrieval scores are retrieved.
In the generation stage, the final answer is generated through the decoding process of LVLMs, based on retrieved contexts.
}

\jh{We observe that current decoding methods adopted by  LVLMs~\cite{lin2022retrieval,lin2024fine,lin2024preflmr,shao2023prompting,lin2022revive} for generation are sub-optimal since they fail to sufficiently leverage relevant contexts, and are disturbed by irrelevant contexts.
In detail, a} decoding method taking only a single context as input significantly degrades as retrieval quality gets worse, as it cannot leverage information from multiple relevant contexts.
On the other hand, while decoding methods taking multiple contexts as input aggregate information from multiple contexts, they are negatively influenced by irrelevant contexts.
Thus, we emphasize the importance of a decoding method that effectively leverages multiple relevant contexts while also successfully handling irrelevant contexts.

In this paper, we propose a novel decoding method for RAG, based on the idea of Contrastive Decoding (CD)~\cite{li2022contrastive,o2023contrastive}, which was initially proposed to enhance the generation results of Large Language Models.
\jh{
CD is an idea of contrasting two different outputs.
It can amplify or counteract the effect of a specific output, which we refer to as \q{reflection} and \q{deflection}.
}
\jh{
Motivated by the idea of CD, we present a simple baseline named Single CD (SCD) for the RAG framework, which amplifies the effect of a context with the highest retrieval score by contrasting the outputs obtained with and without the context.
}
SCD improves RAG by amplifying the context's effect, showing the potential of a reflection in improving RAG with promising retrieval results.
Still, it suffers from significant performance loss when retrieval results get worse, as it cannot leverage multiple relevant contexts.

To this end, we extend the idea of SCD and propose \textbf{R}elevance-aware \textbf{M}ulti-context \textbf{C}ontrastive \textbf{D}ecoding (RMCD), an effective decoding method for RAG that properly reflects and deflects multiple retrieved contexts depending on their relevance to a question.
RMCD outputs a final prediction by combining the predictions obtained with each context.
The weight of each context for combining outputs is adaptively determined based on its retrieval score.
Higher weights are assigned to contexts with high scores, while lower or even negative weights are assigned to contexts with low scores.
Consequently, RMCD outputs the final prediction by reflecting relevant contexts and deflecting irrelevant contexts with high and low retrieval scores.
\jh{Compared to previous decoding methods, RMCD better leverages multiple relevant contexts while also counteracting irrelevant context, thereby generating better responses (Fig.~\ref{fig:moviation_fig}).}

Results on three knowledge-intensive VQA benchmarks, InfoSeek~\cite{chen2023can}, Encyclopedic VQA~\cite{mensink2023encyclopedic}, and OK-VQA~\cite{marino2019ok} show the effectiveness of RMCD as a decoding method for RAG, consistently outperforming other decoding methods applied to \rev{seven} different LVLMs.
RMCD integrates seamlessly into existing LVLMs by simply replacing their decoding methods, \textit{without} additional training.
Moreover, it has the lowest time complexity among methods processing the same number of contexts, achieving a 33\% throughput gain alongside performance improvements (see Table~\ref{tab:decoding_cost_tab}), when applied to the state-of-the-art method on OK-VQA.

Our analyses also demonstrate that RMCD is robust to the results from the retrieval stage.
It performs the best regardless of the retrieval quality, across the weakest to the strongest retrieval results, even including the case where the oracle context is always provided.
Such results show the importance of RMCD, which enhances LVLMs robust to the results from the retrieval stage.

In sum, our contributions are three-fold:
\begin{itemize}
    \item We observe that current decoding methods for Retrieval-augmented Generation (RAG) are insufficiently leveraging multiple relevant contexts and are impacted by irrelevant contexts.
    \item We propose Relevance-aware Multi-context Contrastive Decoding (RMCD), a \textit{training-free} decoding method improving RAG by contrasting multiple contexts based on their relevance to a question, thereby effectively reflecting relevant contexts and deflecting irrelevant contexts.
    \item
    RMCD outperforms other decoding methods on three knowledge-intensive visual question-answering benchmarks when applied to various models, and is also robust to the quality of retrieval results.
\end{itemize}

%% file: tab_fig_tex/motivatoin_fig.tex
\begin{figure}[t!]
\begin{center}
\includegraphics[width=1.0\linewidth]{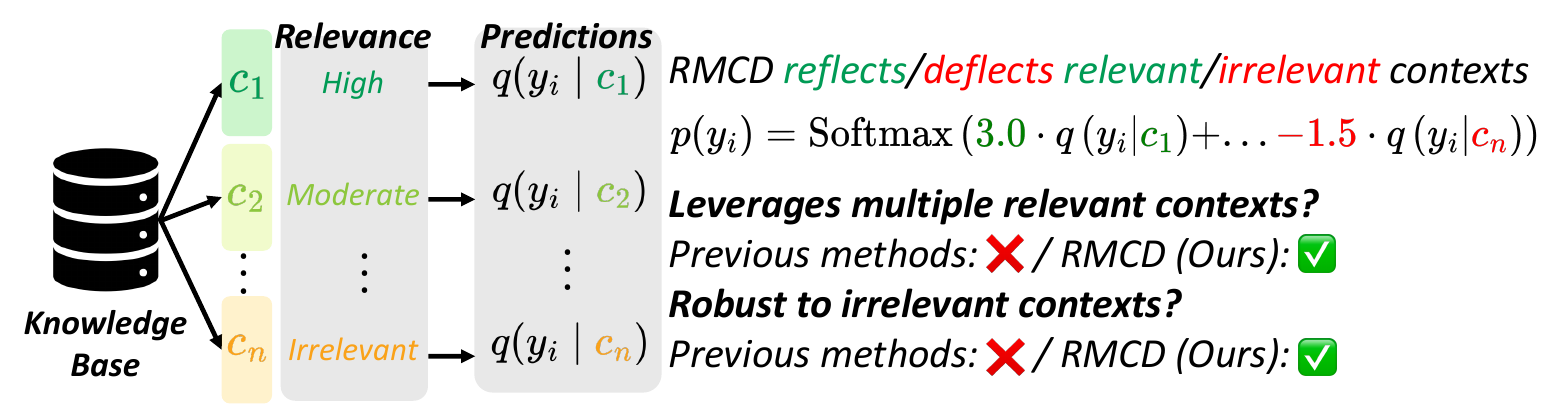}
\end{center}
\vspace{-0.5cm}
\caption{
\textbf{Motivation.}
RMCD generates better responses \textit{without} additional training by contrasting outputs based on relevance of contexts, thereby effectively leveraging multiple {\color{ForestGreen}{relevant}} contexts and counteracting {\color{red}{irrelevant}} contexts.  
}
\label{fig:moviation_fig} 
\vspace{-0.7cm}
\end{figure}

%% file: sec/2_related_works.tex
\section{Related Works}
\subsection{LVLMs and Contrastive Decoding (CD)}
Large Vision Language Models (LVLMs)~\cite{chen2024expanding,li2023blip,dai2024instructblip,liu2024visual,ye2023mplug,zhu2023minigpt,cha2023honeybee,ko2023large} have shown superior performance in various multimodal tasks~\cite{agrawal2019nocaps,goyal2017making,hudson2019gqa,ko2025bidirectional,lee2025vidchain,kim2026tabflash}.
Given a set of input tokens that consist of image features, instructions, and generation results from previous timesteps, LVLMs generate output sequences via autoregressive decoding, which is an iterative process of predicting the next token.
Contrastive Decoding~\cite{li2022contrastive,o2023contrastive} is an improved decoding scheme further enhancing the original prediction result by contrasting it with a prediction from a model with a different size~\cite{li2022contrastive}, from an intermediate layer within the same model~\cite{chuang2024dola}, or from a model conditioned with modulated inputs~\cite{kiminstructive,leng2023mitigating,zhang2023prompt} (\eg perturbed instruction or image).
We extend CD to contrast outputs across multiple contexts, with contrast strength and direction guided by each context’s relevance to the question.

\subsection{Retrieval-augmented Generation (RAG)}
Recently, multiple knowledge-intensive visual question answering benchmarks~\cite{marino2019ok,shah2019kvqa,jain2021select,chen2023can,mensink2023encyclopedic,hu2023open} have been proposed to test models' capability of answering detailed knowledge.
\jh{As a solution, retrieval-augmented visual question answering have been proposed~\cite{gui2021kat,lin2022revive,shao2023prompting,hu2023reveal,lin2022retrieval,lin2024fine,lin2024preflmr}, based on the idea of retrieval-augmented generation~\cite{lewis2020retrieval,park2024generative}, which provides externally retrieved knowledge to models.}
However, effectively utilizing retrieved contexts while preventing the negative effects of those contexts is a crucial challenge as providing irrelevant contexts may even degrade generation quality~\cite{asai2023self,shi2023large,yoranmaking}.
Previous works address this issue with training-based strategies~\cite{gui2021kat,lin2024fine,hu2023reveal,caffagni2024wiki,yoranmaking}, which require a significant computational cost.
In this paper, we propose a novel decoding strategy named Relevance-aware Multi-context Contrastive Decoding (RMCD), enabling better utilization of contexts by contrasting multiple contexts without additional training.

%% file: sec/3_method.tex
\section{Method}
\subsection{Preliminaries}
In this paper, we denote a Large Vision Language Model (LVLM) parameterized by $\theta$ as $\mathcal{M}_{\theta}$.
Decoding of \jh{an} LVLM is an iterative process of generating the most probable sequence through multiple timesteps.
\jh{
At each timestep $i$, \jh{an} LVLM takes image features $I$, instruction $x$ (including a question), and generated sequences from previous timesteps $y_{<i}$ as inputs.
}
Given inputs, \jh{the} LVLM predicts a logit distribution $\mathcal{M}_{\theta}(y_i|I, x, y_{<i})$, which is converted into a probability distribution by softmax as follows:
\begin{equation}
    p(y_i) = \text{Softmax}(\mathcal{M}_{\theta}(y_i | I, x, y_{<i})),
\label{eq:naive_prob}
\end{equation}
where $p(y_i) \in \mathbb{R}^{|\mathcal{V}|}$ is a vector with a dimension of $\mathcal{|V|}$, where $\mathcal{V}$ is a vocabulary set.
In the following sections, we refer to this decoding process as \q{unconditional} decoding 
\jh{as it takes no retrieved contexts as inputs.}

\noindent\textbf{Retrieval-augmented Generation.}
To address knowledge-intensive visual question answering tasks, Retrieval-augmented Generation (RAG)~\cite{lewis2020retrieval} is widely adopted.
RAG complements the model by providing contexts retrieved from an external Knowledge Base (KB) that are relevant to the question.
In the retrieval stage, \jh{textual} contexts \jh{(\eg Wikipedia article)} from the KB that are relevant to the question are retrieved \jh{with off-the-shelf (\eg CLIP~\cite{radford2021learning}) or fine-tuned retrievers~\cite{lin2022retrieval,lin2024preflmr}}.
\jh{
Given an input image $I$ and an instruction $x$ including a question, a query $q$ is set as an image $I$~\cite{radford2021learning} or an image with a question as $q = (I, x)$~\cite{lin2022retrieval,lin2024preflmr}, depending on the retriever. 
Then, a retriever measures the relevance between a query $q$ and every context $c_j$ in KB as:}
\begin{equation}
    s_{c_j} = \text{Retriever}(q; c_j).
\end{equation}
The retrieved context set $\mathcal{C}$ is then defined, which consists of $n$ contexts with the highest retrieval scores, where elements in the set are sorted in descending order by their scores.
After the retrieval stage, the generation stage follows, where an LVLM generates a response considering retrieved contexts.
The decoding process is the process of generating the final probability distribution with retrieved contexts $\mathcal{C}$, in which those are fed into an LVLM as additional inputs as:
\begin{equation}
    p(y_i) = \text{Softmax}(\mathcal{M}_{\theta}(y_i | I, x, y_{<i}, \mathcal{C})),
\label{eq:naive_rag_prob}
\end{equation}
where information from a part or whole context in $\mathcal{C}$ is aggregated, depending on the decoding method.

\input{tab_fig_tex/main_fig}
\subsection{Baseline: Single-context Contrastive Decoding}
\rev{Here}, we present \jh{a} simple adoption of CD into the decoding process of RAG.
\rev{We define} a simple decoding strategy that contrasts the difference between an unconditional logit and a logit conditioned on a context $c_1$, as follows:
\begin{equation}
\begin{split}
     p(y_i) =  
     \text{Softmax}(&\alpha_1 \mathcal{M}_{\theta}(y_i | I, x, y_{<i}, c_1) \\
     &- \alpha_2 \mathcal{M}_{\theta}(y_i | I, x, y_{<i})),
\end{split}
\label{eq:rag_cd_prob}
\end{equation}
where $\alpha_1$ and $\alpha_2$ are hyperparameters, and $c_1 \in \mathcal{C}$ is a retrieved context with the highest retrieval score.
We refer to the decoding strategy as Single-Context Contrastive Decoding (SCD), as it leverages a difference between an unconditional logit and a single logit with a context.
Through experiments, we observe that SCD outperforms most of the previous decoding methods when retrieval results are promising, as actively reflecting contexts is largely helpful in such cases (Tab.~\ref{tab:main_table}).
\rev{However, we also observe that SCD significantly degrades as the retrieval quality gets worse (Fig.~\ref{fig:pool_size_performance_fig}).
This limitation shows that SCD cannot effectively handle noisy retrieval, as it relies on a fixed contrast strength with only one context rather than dynamically leveraging multiple contexts.}

\subsection{Relevance-aware Multi-context Contrastive Decoding (RMCD)}
\rev{To this end, we propose a \textbf{R}elevance-aware \textbf{M}ulti-context \textbf{C}ontrastive \textbf{D}ecoding (RMCD), a training-free method effectively leveraging multiple contexts by adaptively contrasting multiple logits obtained for each context based on their relevance to a question.
}
To do so, context weights, which determine the degree of reflection and deflection of each context, are firstly obtained based on their retrieval scores (Eq.~\eqref{eq:multi_cd_coef}).
Then, the final prediction is defined as a combination of logits, where each logit is weighted by its corresponding context weight (Eq.~\eqref{eq:multi_cd}).
Consequently, RMCD more effectively reflects the influence of relevant contexts, while deflecting the effect of irrelevant contexts.
The overall pipeline of RMCD is illustrated in Fig.~\ref{fig:main_fig}.

\noindent\textbf{Relevance-aware multi-context contrastive decoding.}
To predict a probability distribution at timestep $i$, RMCD utilizes contexts from the retrieved context set $\mathcal{C}$, where contexts $c_j \in \mathcal{C}$ are sorted in descending order by the retrieval score $s_{c_j}$.
Note that RMCD includes an empty context $c_{n+1}$ with $s_{c_{n+1}}$ set to $-\infty$ in $\mathcal{C}$, which corresponds to an unconditional generation.
In the supplementary material, we report results with other design choices in determining $s_{c_{n+1}}$, while we empirically found $-\infty$ worked the best.
For every context $c_j \in \mathcal{C}$, a predicted logit is first obtained as:
\begin{equation}
    q(y_i|c_j) = \mathcal{M}_{\theta}(y_i|I, x, y_{<i}, c_j),
\label{eq:context_logit}
\end{equation}
where $q(y_i|c_j) \in \mathbb{R}^{|\mathcal{V}|}$ is a vector of a predicted logit distribution provided a context $c_j$.
Note that inputs other than the context $c_j$ in $q(\cdot)$ are omitted for conciseness in the rest of the paper.
Then, the final prediction logit in RMCD is defined as a weighted sum of predicted logits $q(y_i|c_j)$, with each logit weighted by a context weight $\alpha_j$.
\jh{
Here, we introduce the process of determining $\alpha_j$ of each context $c_j$ using retrieval score $s_{c_j}$.
}
To calculate $\alpha_j$, we first calculate relative context scores $w_{c_j}$ using $s_{c_j}$ as below:
\begin{equation}
    w_{c_j} = \frac{\exp(s_{c_j} / \tau_1)}{\sum_{k=1}^{|\mathcal{C}|}\exp(s_{c_k} / \tau_1)},
\label{eq:coef_softmax}
\end{equation}
where $\tau_1$ is a hyperparameter determining the softmax temperature.
Calculated relative context score $w_{c_j}$ implies the relative relevance of a context compared to other retrieved contexts in $\mathcal{C}$. 
\jh{
Then, based on $w_{c_j}$, the context weight $\alpha_j$ is obtained with the mapping process below.
Note that the transformation below allows $\alpha_j$ to have a negative value, thus enabling deflection of information from contexts with low scores.
The process is defined as:
}
\begin{equation}
    \alpha_j = M - \delta\left(\frac{w_{c_1} - w_{c_j}}{w_{c_1}}\right),
\label{eq:multi_cd_coef}
\end{equation}
where $\delta = (M - m)$ is a gap between two hyperparameters, $M$ and $m$, which determine the maximum and minimum context weights, respectively.
\jh{Note that $m$ is set as a negative value to assign negative weights to $\alpha_j$.}
Based on the formula above, $\alpha_j$ is defined with a ratio between $w_{c_j}$ and $w_{c_1}$, which are relative scores of the context $c_j$ and the most likely context $c_1$.
If $w_{c_j}$ is close to $w_{c_1}$, $\alpha_j$ is set close to $M$.
On the contrary, if $w_{c_j}$ is significantly smaller than $w_{c_1}$, $\alpha_j$ is set close to $m$.
Then, softmax is applied to the weighted sum of logits to form a predicted probability distribution at the timestep $i$.
The decoding process of RMCD is formally defined as:
\begin{equation}
    p(y_i) = \text{Softmax}(\sum_{j=1}^{|\mathcal{C}|} \alpha_j q(y_i|c_j)).
\label{eq:multi_cd}
\end{equation}
With RMCD, a larger weight is applied to a prediction logit with relatively higher retrieval scores. 
In comparison, a smaller or even negative weight is applied to a logit with lower retrieval scores, actively reflecting relevant contexts and deflecting irrelevant contexts' influence.

\noindent\textbf{Ensembled plausibility constraint.}
Although RMCD helps effective utilization of contexts by adjusting the probabilities of vocabularies to be sampled, sampling from \jh{the} entire vocabulary $v \in \mathcal{V}$ may be sub-optimal, as shown in~\cite{li2022contrastive}, since some vocabularies are highly unlikely to be generated.
Thus, we develop an ensembled plausibility constraint restricting the plausible vocabulary set to sample from, based on ensembled predictions of constraint contexts, expanding the idea of a plausibility constraint~\cite{li2022contrastive}.
To apply the ensembled plausibility constraint, we first obtain constrained logits with $c_j \in \mathcal{C}$ as below:
\begin{equation}
    \tilde{q}(y_i|c_j)_v = 
    \begin{cases}
        q(y_i|c_j)_v & \text{if } v\in \mathcal{S}_i, \\
        -\infty & \text{if } v \notin \mathcal{S}_i,
    \end{cases}
\end{equation}
where $\mathcal{S}_i \in \mathcal{V}$ is a subset of the whole vocabulary set $\mathcal{V}$ consisting only of plausible vocabularies at $i$-th timestep, in which the detailed process to obtain $\mathcal{S}_i$ will be elaborated shortly after.
Given an original logit $q(y_i|c_j)$, a constrained logit $\tilde{q}(y_i|c_j)$ is obtained by replacing the $v$-th entry of the logit with $-\infty$ if the corresponding vocabulary is not in the plausible set $\mathcal{S}_i$.
Then, the final decoding process of RMCD with constraints is defined as:
\begin{equation}
    p(y_i) = \text{Softmax}(\sum_{j=1}^{|\mathcal{C}|} \alpha_j \tilde{q}(y_i|c_j)),
    \label{eq:mcd_with_constraint}
\end{equation}
where the decoding process is identical to Eq.~\eqref{eq:multi_cd}, except that the original logit $q(y_i|c_j)$ is replaced with the constrained logit $\tilde{q}(y_i|c_j)$.
With constraints, only the vocabularies in $\mathcal{S}_i$ are allowed to be sampled, since the probability is set to zero otherwise.

From now on, we elaborate on the detailed process to obtain $\mathcal{S}_i$.
First, we define a constraint context set $\mathcal{C}^c := \{c_k \in \mathcal{C} \ | \ w_{c_k} \geq \gamma\}$, which is a subset of $\mathcal{C}$ consisting of contexts with a relative context score $w_{c_k}$ larger than the score threshold $\gamma$.
Then, the relative importance of each constraint context is calculated as:
\begin{equation}
    \tilde{w}_{c_k} = \frac{\exp(s_{c_k} / \tau_2)}{\sum_{l = 1}^{|\mathcal{C}^c|}\exp(s_{c_l} / \tau_2)},
\label{eq:promising_coef_softmax}
\end{equation}
where $\tau_2$ is a hyperparameter for the softmax temperature.
Note that the softmax is applied only to contexts in the constraint context set $\mathcal{C}^c$ rather than the whole retrieved context set $\mathcal{C}$.
\jh{The ensembled} probability $\tilde{p}(y_i)$ is then defined as a weighted sum of predicted logits given a constraint context $c_k \in \mathcal{C}^c$, with each logit weighted by $\tilde{w}_{c_k}$, followed by a softmax as:
\begin{equation}
\tilde{p}(y_i) = \text{Softmax}(\sum_{k = 1}^{|\mathcal{C}^c|}\tilde{w}_{c_k} q(y_i|c_k)).
\end{equation}
Based on the ensembled probability, the set $\mathcal{S}_i$ of plausible vocabs at timestep $i$ is defined as follows:
\begin{equation}
    \mathcal{S}_i := \{v \in \mathcal{V}  \ |\  \beta \max_t \tilde{p}(t) \leq \tilde{p}(v)\},
    \label{eq:plausible_vocab_set}
\end{equation}
where $\beta \in (0, 1)$ is a hyperparameter for the constraint strength.
With Eq.~\eqref{eq:plausible_vocab_set}, only vocabularies with probabilities above a certain threshold relative to the highest probability are included in $\mathcal{S}_i$.
We provide ablation results in Tab.~\ref{tab:constraint_set_ablation_tab}, showing the effectiveness of threshold-based construction of $\mathcal{S}_{i}$.

%% file: tab_fig_tex/main_fig.tex
\begin{figure*}[t!]
\begin{center}
\includegraphics[width=1.0\textwidth]{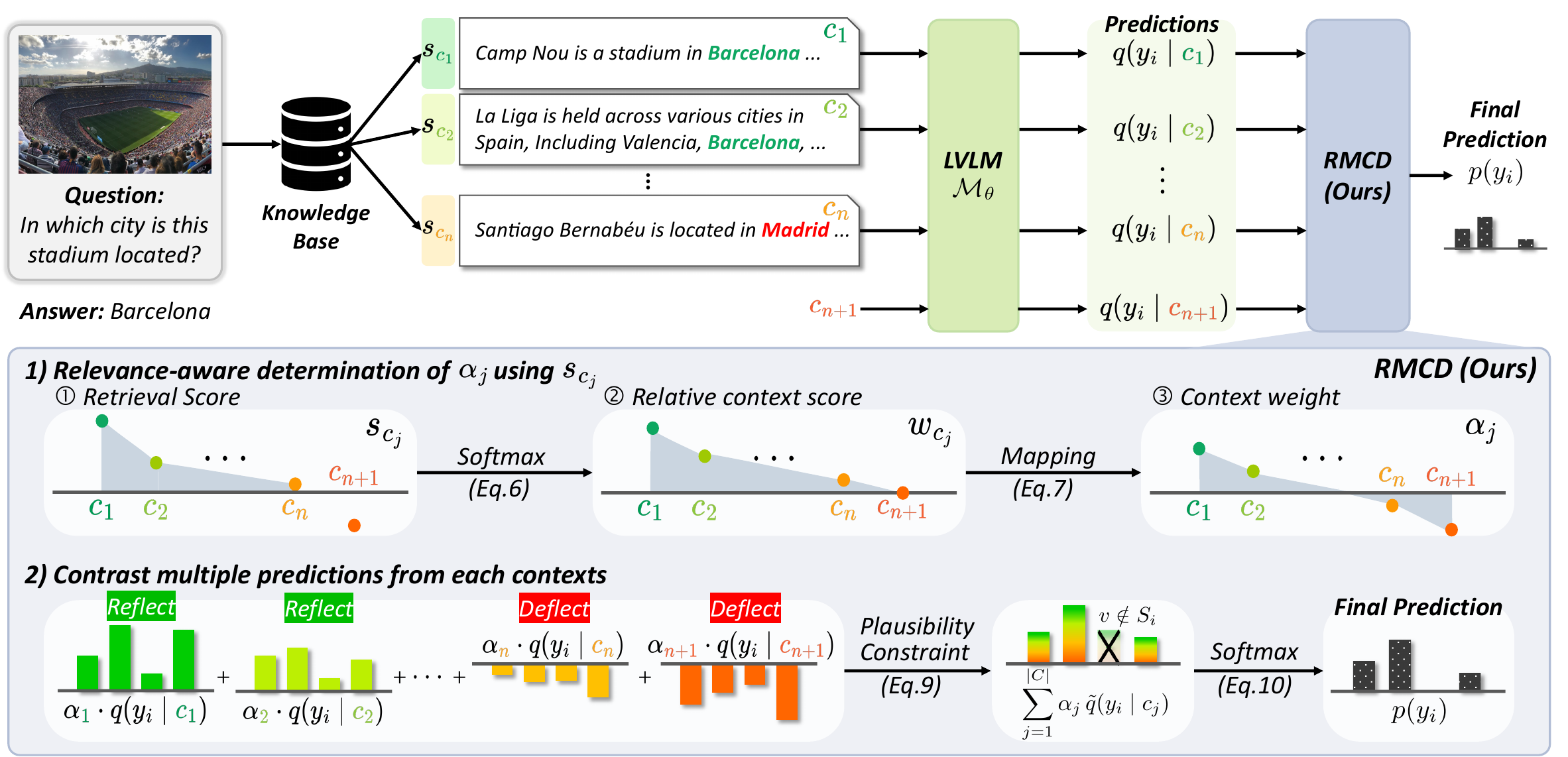}
\end{center}
\vspace{-0.4cm}
\caption{
\textbf{Relevance-aware Multi-context Contrastive Decoding (RMCD).}
An LVLM $\mathcal{M}_{\theta}$ outputs multiple predictions $q(y_i| c_j)$ provided contexts $c_{j} \in \mathcal{C}^r$ retrieved from a knowledge base.
Using the retrieval score $s_{c_j}$ of a context, a context weight $\alpha_j$ is obtained.
Then, RMCD combines multiple predictions by contrasting multiple contexts, where each prediction is weighted by $\alpha_j$.
Thereby, outputs with high $\alpha_j$ are effectively {\color{ForestGreen}{reflected}} while outputs with low $\alpha_j$ are {\color{Red}{deflected}} as illustrated in the figure.
}
\label{fig:main_fig}
\vspace{-0.55cm}
\end{figure*}

%% file: sec/4_experiments.tex
\section{Experiments}
In this section, we report results on three knowledge-intensive VQA benchmarks, InfoSeek~\cite{chen2023can}, Encyclopedic-VQA~\cite{mensink2023encyclopedic}, and OK-VQA~\cite{marino2019ok} by applying RMCD to multiple LVLMs.
Further details about baseline LVLMs, benchmarks, and metrics are in the supplementary material.

\subsection{Datasets and metrics}
\noindent\textbf{InfoSeek}~\cite{chen2023can} is a knowledge-intensive VQA benchmark.
We report the performance on the validation set, which consists of 73.6k samples.
Following~\cite{chen2023can}, the total accuracy is reported as a metric, defined as a harmonic mean of accuracy on two subsets: unseen question and unseen entity.
\textbf{Encyclopedic-VQA}~\cite{mensink2023encyclopedic} is another knowledge-intensive VQA benchmark.
We report the performance on the test set, which consists of 5.7k samples.
Questions are categorized into single-hop, two-hop, and multi-answer types.
Following the original paper~\cite{mensink2023encyclopedic}, we report the average accuracy on single-hop questions (\ie one-hop) as a main result, while we also provide the average accuracy on all samples (\ie full).
\textbf{OK-VQA}~\cite{marino2019ok} is also a knowledge-intensive VQA benchmark.
We report the VQA score on the test set \jh{containing} 5K samples, following~\cite{marino2019ok,lin2024fine,hu2023reveal}.
\input{tab_fig_tex/new_main_tab}
\input{tab_fig_tex/okvqa_tab}

\input{tab_fig_tex/cost_tab}

\subsection{Implementation details}
\textbf{Base LVLMs.}
For InfoSeek and Encyclopedic-VQA benchmarks, we report performances of RMCD on \rev{five} variants across \rev{three} different LVLMs, \rev{InternVL-2.5~\cite{chen2024expanding},} BLIP-2~\cite{li2023blip}, and LLaVA-1.5~\cite{liu2023improved}.
\rev{As a base language model (base LM), Qwen2.5-32B~\cite{yang2024qwen2} is applied for InternVL-2.5 (38B)}, T5-XXL~\cite{chung2024scaling}, T5-XL, and OPT 6.7~\cite{zhang2022opt} are applied for BLIP-2, and Vicuna-13B~\cite{vicuna2023} is applied for LLaVA-1.5.
For OK-VQA, we implement RMCD on top of RA-VQA~\cite{lin2022retrieval} and FLMR~\cite{lin2024fine}, competitive architectures with the RAG pipeline \jh{adopting} T5-XL as a base LM.
All experiments are done with 4 NVIDIA RTX A6000 GPUs.

\noindent\textbf{Retrieval procedure.}
For \textbf{InfoSeek}, we constructed a Knowledge Base (KB) from a Wikipedia dump following the original paper, as the original KB was unavailable. 
Each KB entry includes a Wikipedia article, its summary, and an associated entity image. 
We detail the KB construction in supplementary materials. 
Retrieval for InfoSeek uses CLIP~\cite{radford2021learning}.
For \textbf{Encyclopedic-VQA}, we employ the original Wikipedia KB and primarily use retrieval results from Google Lens~\cite{googlelens}, provided by the authors. 
When fewer than five contexts are retrieved via Google Lens, we supplement additional contexts using CLIP. 
Retrieval details for both datasets are in the supplementary materials.
The details of the retrieval process on InfoSeek and Encyclopedic-VQA are in the supplementary material.
For \textbf{OK-VQA}, we directly adopt retrieval results from baseline models RA-VQA~\cite{lin2022retrieval} and FLMR~\cite{lin2024fine}, using contexts retrieved from Google Search Corpus~\cite{luo2021weakly} with their respective retrievers.

\noindent\textbf{Hyperparameters.}
We adopt $\tau_1 = 1.75, \tau_2 = 0.5, \gamma = 0.3, M = 4, m = -1, \beta=0.2$ as default for InfoSeek benchmark.
We only adjust $\tau_1 = 3.0$ for Encyclopedic-VQA, and $M = 3.0$ and $ \tau_1=3.5$ for OK-VQA while maintaining every other value the same.
Note that the same hyperparameters are applied across LVLMs within the same dataset.
In the supplementary material, we provide a sensitivity analysis showing that the performance of RMCD is robust to hyperparameters, consistently outperforming other methods regardless of hyperparameter choice.

\subsection{Baselines}
\textbf{Decoding methods.} We provide details about six decoding methods as baselines: unconditional, RAG, SCD, consistency, max probability, and concat.
\q{\textbf{Unconditional}} (Eq.~\ref{eq:naive_prob}) and \q{\textbf{RAG} $(n=1)$} (Eq.~\ref{eq:naive_rag_prob}) denote methods that take no context and a single context $c_1$ with the highest retrieval score as input, respectively.
For simplicity, we omit $(n=1)$ in the rest of the paper when referring to RAG $(n=1)$ as a decoding method.
\q{\textbf{SCD}} (Eq.~\ref{eq:rag_cd_prob}) is similar to RAG, but the effect of the context is amplified, where $(\alpha_1, \alpha_2)$ is set as $(2, 1)$.
As methods taking multiple contexts as inputs, we adopt \q{consistency}, \q{max probability}, and \q{concat}.
For \q{\textbf{consistency}} and \q{\textbf{max probability}}, $n$ predictions are generated independently given $n$ retrieved contexts. 
Among these predictions, \q{consistency} selects the most frequent answer, similar to~\cite{wang2022self}, while \q{max probability} selects the answer with the highest average confidence over the generated tokens~\cite{lin2022retrieval,lin2024fine,lin2024preflmr}.
In cases where multiple answers tie, an answer is randomly chosen from the tied answers.
For \q{\textbf{concat}}, $n$ retrieved contexts are concatenated into a single string and provided to an LVLM as an input context to generate a single answer~\cite{gui2021kat,lin2022revive,shao2023prompting,hu2023reveal}.
\jh{
Unless specified, $n=5$ is applied for methods taking multiple contexts as input, following recent works~\cite{lin2022retrieval,lin2024fine}.
}

\subsection{Results on InfoSeek and Encyclopedic-VQA}
In Tab.~\ref{tab:main_table}, performances of decoding methods are reported on InfoSeek and Encyclopedic-VQA benchmarks.
\jh{The basic version proposed, SCD outperforms majority of other decoding methods, while maintaining the efficiency.
RMCD further improves SCD by taking benefits of multiple contexts, thereby achieving the best results of \rev{22.9, 41.4, and 34.6} on InfoSeek, Encyclopedic-VQA one-hop, and full splits.
}
Compared to the unconditional generation without the retrieval, RMCD improves performance on InfoSeek by up to 13.1 points and also shows improvements by up to 24.5 and 18.3 points on one-hop and full split of Encyclopedic-VQA.
While every other method underperforms the simplest RAG on InfoSeek when applied to BLIP-2 (T5-XL), RMCD improves RAG by 0.7 points, showing its robustness.
Overall, the results show that RMCD is the most effective decoding method showing the best performance across multiple LVLMs and benchmarks\jh{, while SCD, the simple variant of RMCD, is also a cost-efficient choice that still outperforms majority of other methods}.

\subsection{Comparison with state-of-the-art on OK-VQA}
\input{tab_fig_tex/compare_with_multi_context_fig}
The results on OK-VQA are reported in Tab.~\ref{tab:ok_vqa_tab}, where we implement RMCD on the two competitive architectures, RA-VQA~\cite{lin2022retrieval} and FLMR~\cite{lin2024fine} by replacing their decoding methods with RMCD.
As a result, improvements of 0.7 and 1.0 points in accuracy are achieved, resulting in the best accuracy of 63.1 with FLMR + RMCD.
RMCD also outperforms methods that use GPT-3 in the retrieval~\cite{gui2021kat, lin2022revive} or generation~\cite{shao2023prompting} stage, requiring a huge amount of computation.
Also, unlike every other method that trains a model for VQA tasks, RMCD does \textit{not} require additional training.
Results show that RMCD is also applicable to architectures extensively developed toward a specific benchmark to improve them, showing the \jh{expandability} of RMCD.
Regarding decoding costs, RMCD achieves the lowest time complexity, approximately $\mathcal{O}(n)$ relative to the number of contexts $n$, since it avoids techniques like beam search or ensembling.
As a result, replacing FLMR's decoding method (max probability + beam search) which has a \jh{complexity} of $2 \cdot \mathcal{O}(n)$ with RMCD reduces the latency by 33\%  (1.55 to 1.04 \jh{sec.} per sample, Tab.~\ref{tab:decoding_cost_tab}).
Similarly, replacing RA-VQA's decoding method (max probability) having the same complexity of $\mathcal{O}(n)$ with RMCD only induces 4.4\% loss in latency (Tab.~\ref{tab:decoding_cost_tab}), showing a minimal computation overhead of RMCD.
Note that we only compared the latency difference caused by replacing baselines' 
decoding with RMCD, as methods in Tab.~\ref{tab:ok_vqa_tab} vary significantly in architecture and size, making direct comparison challenging.

%% file: tab_fig_tex/new_main_tab.tex
{\renewcommand{\arraystretch}{1.1}
\begin{table*}[t!]
\centering
    \begin{adjustbox}{width=\textwidth}
    \begin{tabular}{l|*{4}{ccc|}ccc}
        \toprule
         \multirow{3}{*}{\vspace{-2mm}Decoding Method}& 
         \multicolumn{3}{c|}{\textbf{\rev{InternVL 2.5 (38B)~\cite{chen2024expanding}}}} &
         \multicolumn{3}{c|}{\textbf{BLIP-2 (T5-XXL)~\cite{li2023blip}}} & 
         \multicolumn{3}{c|}{\textbf{BLIP-2 (T5-XL)~\cite{li2023blip}}} & 
         \multicolumn{3}{c|}{\textbf{LLaVA-1.5 (Vicuna 13B)~\cite{liu2023improved}}} & 
         \multicolumn{3}{c}{\textbf{BLIP-2 (OPT 6.7B)~\cite{li2023blip}}} \\
         \cmidrule{2-16}
        & \textbf{\rev{InfoSeek}} & \multicolumn{2}{c|}{\textbf{\rev{E-VQA}}} 
        & \textbf{InfoSeek} & \multicolumn{2}{c|}{\textbf{E-VQA}} 
        & \textbf{InfoSeek} & \multicolumn{2}{c|}{\textbf{E-VQA}} 
        & \textbf{InfoSeek} & \multicolumn{2}{c|}{\textbf{E-VQA}}
        & \textbf{InfoSeek} & \multicolumn{2}{c}{\textbf{E-VQA}} \\
        & \rev{Full} & \rev{One-hop} & \rev{Full} & Full & One-hop & Full & Full & One-hop & Full & Full & One-hop & Full & Full & One-hop & Full \\
        \midrule
        Unconditional   & \rev{13.3} & \rev{19.8} & \rev{17.0} & 7.9 & 15.0 & 12.7 & 7.3 & 13.3 & 10.6 & 7.7 & 13.6 & 12.6 & 4.0 & 10.8 & 7.7 \\
        RAG             & \rev{20.7} & \rev{37.5} & \rev{31.1} & 19.3 & 35.0 & 27.9 & 17.4 & 34.5 & 27.4 & 18.9 & 23.8 & 18.5 & 4.0 & 14.3 & 10.6 \\
        SCD             & \rev{20.4} & \rev{37.9} & \rev{32.2} & 19.6 & 36.0 & 29.0 & 17.1 & 34.6 & 27.4 & \textbf{19.6} & 25.7 & 20.9 & 4.9 & 17.2 & 13.1 \\
        Consistency     & \rev{13.8} & \rev{22.6} & \rev{24.6} & 12.0 & 24.4 & 12.2 & 9.6 & 21.4 & 17.6 & 12.0 & 19.9 & 16.3 & 1.2 & 9.6 & 6.3 \\
        Concat          & \rev{22.3} & \rev{40.9} & \rev{33.9} & 18.7 & 37.4 & 30.9 & 12.1 & 34.9 & 26.0 & 14.9 & 22.6 & 17.6 & 0.5 & 5.1 & 6.0 \\
        Max Probability & \rev{17.2} & \rev{23.3} & \rev{34.1} & 15.5 & 36.6 & 12.4 & 14.8 & 35.4 & 27.2 & 16.9 & \textbf{27.0} & 21.5 & 1.2 & 14.9 & 10.6 \\
        \midrule
        \rowcolor{lightblue}
        RMCD \textit{(Ours)} 
                        & \rev{\textbf{22.9}} & \rev{\textbf{41.4}} & \rev{\textbf{34.6}}
                        & \textbf{21.0} & \textbf{39.5} & \textbf{31.0} 
                        & \textbf{18.1} & \textbf{36.3} & \textbf{28.6} 
                        & 19.2 & \textbf{27.0} & \textbf{22.1} 
                        & \textbf{5.8} & \textbf{19.0} & \textbf{14.1} \\
        \bottomrule
    \end{tabular}
    \end{adjustbox}
    \vspace{-0.3cm}
    \caption{\textbf{Performance on InfoSeek~\cite{chen2023can} and Encyclopedic-VQA~\cite{mensink2023encyclopedic} by decoding method.} Best results on each benchmark among the same baseline LVLM are marked \textbf{bold}.}
    \label{tab:main_table}
    \vspace{-0.6cm}
\end{table*}
}

%% file: tab_fig_tex/okvqa_tab.tex
\begin{table}[t!]
\centering
\begin{adjustbox}{width=\linewidth}
\begin{tabular}{l|ccc|c}
\toprule
Method & Decoding Cost & Train & GPT-3 & Acc. \\
\midrule
KAT~\cite{gui2021kat} & $3 \cdot \mathcal{O}(n)$ &\cmark & \cmark & 54.4 \\
REVIVE~\cite{lin2022revive}& $3 \cdot \mathcal{O}(n)$ & \cmark & \cmark & 58.0 \\
Prophet~\cite{shao2023prompting} & $5 \cdot \mathcal{O}(n^2)$ & \cmark & \cmark & 61.1 \\
ReVeaL~\cite{hu2023reveal} & $\mathcal{O}(n)$ & \cmark & \xmark &59.1 \\
PreFLMR~\cite{lin2024preflmr} & $2 \cdot \mathcal{O}(n)$ & \cmark & \xmark &61.9 \\
\midrule
RA-VQA~\cite{lin2022retrieval} & $\mathcal{O}(n)$ & \cmark & \xmark &51.2 \\
\rowcolor{lightblue}
RA-VQA + RMCD \textit{(Ours)} &$\mathcal{O}(n)$ & \xmark & \xmark & 51.9\\
FLMR~\cite{lin2024fine} & $2 \cdot \mathcal{O}(n)$ & \cmark & \xmark &62.1 \\
\rowcolor{lightblue}
FLMR + RMCD \textit{(Ours)}& $\mathcal{O}(n)$ & \xmark & \xmark & \textbf{63.1} \\
\bottomrule
\end{tabular}
\end{adjustbox}
\vspace{-0.3cm}
\caption{\textbf{Performance on OK-VQA~\cite{marino2019ok}.} $n$: \# retrieved contexts.
Train: Require additional training, GPT-3: Use GPT-3~\cite{brown2020language}, Acc: VQA Score~\cite{marino2019ok}. Coefficients in decoding cost denote \# model forwarding for ensemble (KAT, REVIVE, Prophet) and beam width (FLMR, PreFLMR). The best result is marked \textbf{bold}.}
\label{tab:ok_vqa_tab}
\vspace{-0.5cm}
\end{table}


%% file: tab_fig_tex/cost_tab.tex
\begin{table}[t!]
\centering
    \centering
    \begin{adjustbox}{width=\linewidth}
    \begin{tabular}{l|ccc}
        \toprule
        Method & Decoding Cost & Latency $\downarrow$ & Accuracy $\uparrow$ \\
        \midrule
        RA-VQA~\cite{lin2022retrieval} & $\mathcal{O}(n)$ & 0.45 & 51.2 \\
        \rowcolor{lightblue}
        RA-VQA + RMCD \textit{(Ours)} & $\mathcal{O}(n)$ & 0.47 \small (\color{Gray}{$\uparrow$ 4.4\%})& \textbf{51.9} \small \color{ForestGreen} {({$\uparrow$ 0.7\%p})}\\
        FLMR~\cite{lin2024fine} & $2 \cdot \mathcal{O}(n)$ & 1.55 & 62.1 \\
        \rowcolor{lightblue}
        FLMR + RMCD \textit{(Ours)} & $\mathcal{O}(n)$ & 1.04 \small (\color{ForestGreen}{$\downarrow$ 32.9\%})& \textbf{63.1} \small \color{ForestGreen}{({$\uparrow$ 1.0\%p})} \\
        \bottomrule
    \end{tabular}
    \end{adjustbox}
    \vspace{-0.3cm}
    \caption{\textbf{Decoding cost comparison.} Latency: seconds per sample. The coefficient on the cost of FLMR denotes beam width.}
    \vspace{-0.8cm}
    \label{tab:decoding_cost_tab}
\end{table}

%% file: tab_fig_tex/compare_with_multi_context_fig.tex
\begin{figure*}[ht!]
\centering
\begin{minipage}{0.32\textwidth}
    \input{tab_fig_tex/n_contexts_performance_plot}
\end{minipage}
\hfill
\begin{minipage}{0.32\textwidth}
    \input{tab_fig_tex/pool_size_recall_fig}
\end{minipage}
\begin{minipage}{0.32\textwidth}
    \input{tab_fig_tex/pool_size_performance_fig}
\end{minipage}
\vspace{-0.65cm}
\end{figure*}

%% file: tab_fig_tex/n_contexts_performance_plot.tex

\centering
\includegraphics[width=0.95\textwidth]{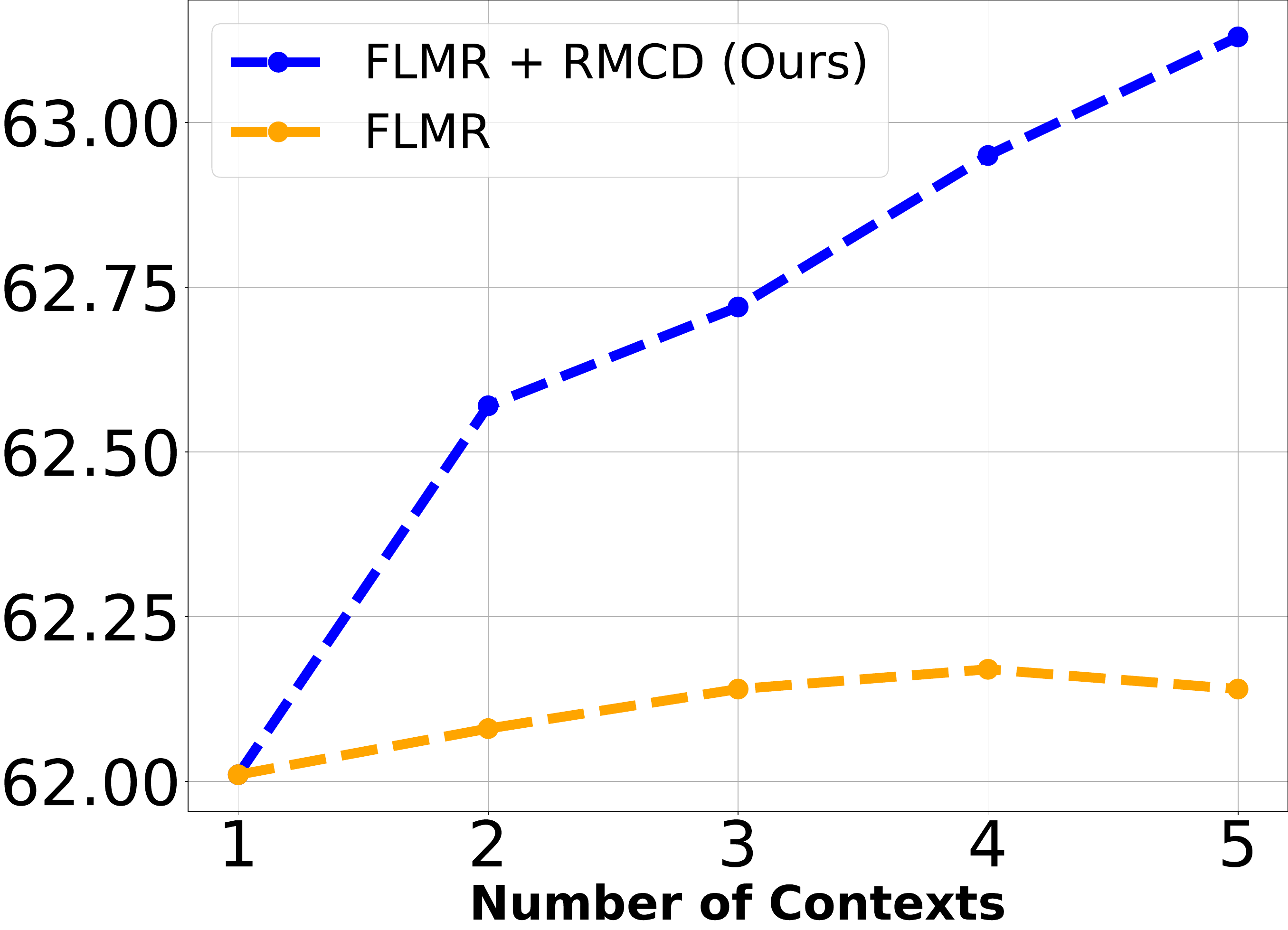}
\vspace{-0.2cm}
\caption{
\small \textbf{Accuracy by the number of contexts $n$ used.}
}
\label{fig:n_contexts_performance_fig}

%% file: tab_fig_tex/pool_size_recall_fig.tex
\centering
\includegraphics[width=0.95\textwidth]{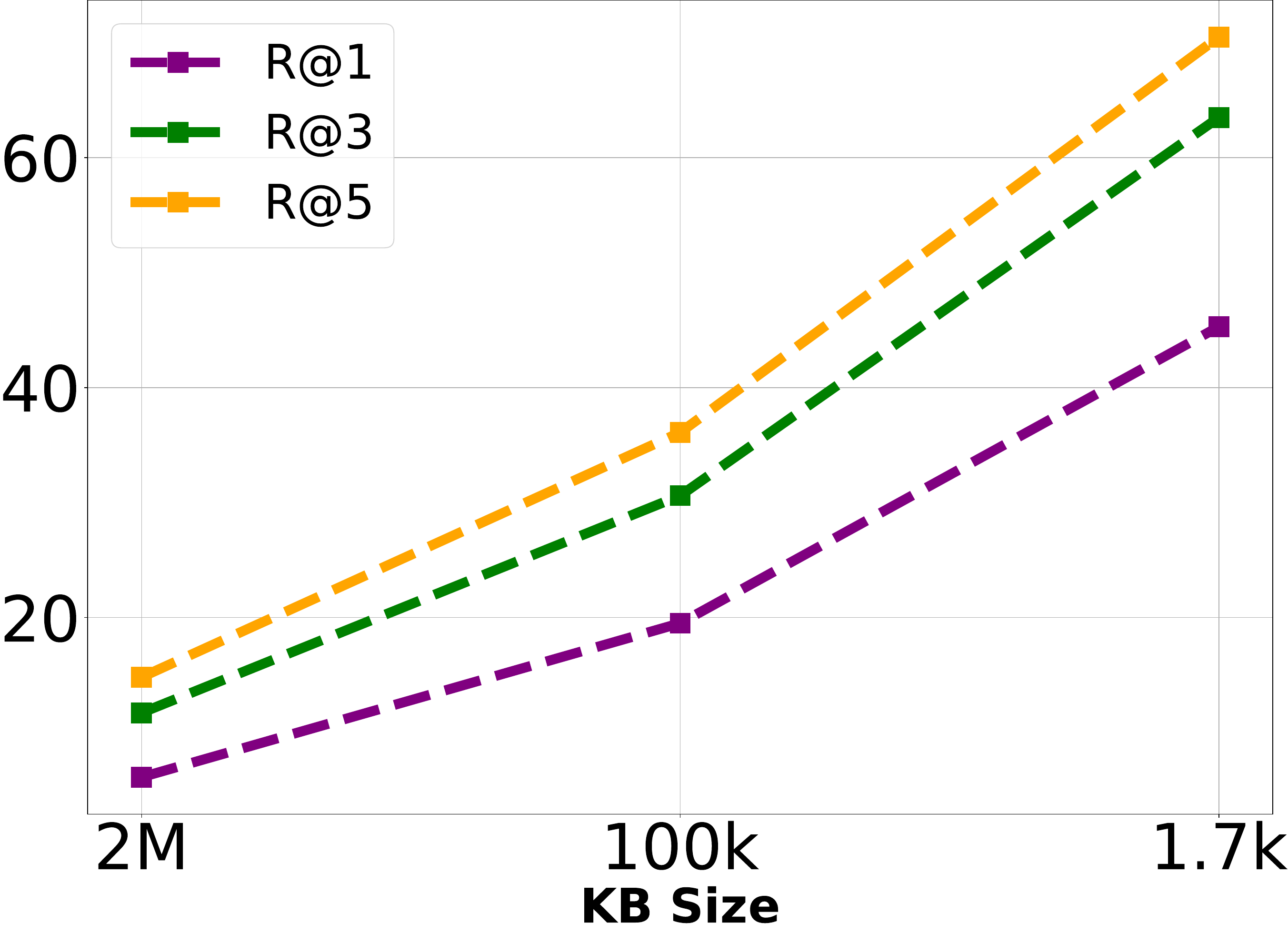}
\vspace{-0.15cm}
\caption{
\small \textbf{Retrieval Recall@$k$ ($k=1, 3, 5$) by the size of KBs.}
}
\label{fig:pool_size_recall_fig}

%% file: tab_fig_tex/pool_size_performance_fig.tex
\centering
\includegraphics[width=0.95\textwidth]{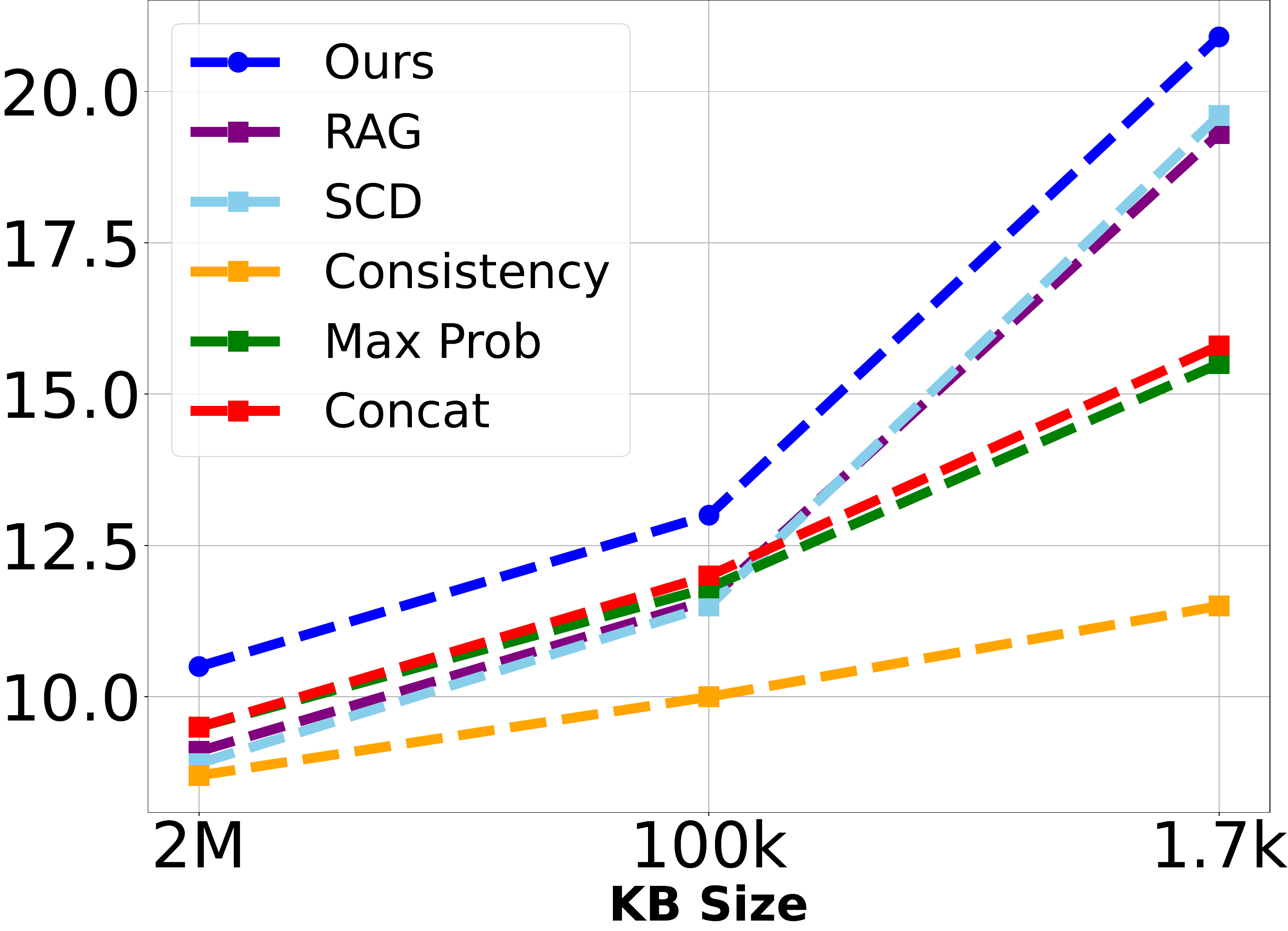}
\vspace{-0.15cm}
\caption{
\small \textbf{Accuracy by the size of KBs. Retrieval gets more difficult in larger KBs.}
}
\label{fig:pool_size_performance_fig}

%% file: sec/5_analysis.tex
\section{Analysis}
\jh{Here, we provide the performance of decoding methods on InfoSeek, when applied to BLIP-2 (T5-XXL) for analysis.
\subsection{Effect of the quantity of retrieved contexts}}
\label{sec:analysis}
Performance of FLMR, a state-of-the-art architecture for RAG, with baseline decoding method and RMCD across different numbers of contexts $n$ is illustrated in Fig.~\ref{fig:n_contexts_performance_fig}. 
FLMR + RMCD consistently benefits from leveraging more contexts, exhibiting a monotonic increase in performance as $n$ increases. 
In contrast, the baseline struggles to effectively utilize additional contexts, showing negligible improvement or even performance degradation. 
Consequently, the performance gap between FLMR + RMCD and the baseline widens progressively, reaching a notable difference of 1\%p at $n = 5$. 
This result highlights RMCD's superior capability in effectively leveraging multiple contexts.

\jh{\subsection{Effect of the quality of retrieved contexts}
We conduct two controlled experiments to answer: \textit{Why use RMCD instead of just improving the retriever?}
Results show RMCD consistently outperforms all methods across retriever strengths, including the oracle retriever, proving its complementary value beyond retriever improvements.
}

\noindent\textbf{Performance with varying retriever capabilities.}
\jh{
To simulate retrievers of varying capabilities while keeping other factors constant, we constructed KBs of three sizes: 1.7K, 100K, and 2M. 
The 1.7K KB, used in all other experiments, contains only contexts relevant to questions in InfoSeek benchmark. 
Adding irrelevant contexts expands it to 100K and 2M, reducing retrieval performance in larger KBs (Fig.~\ref{fig:pool_size_recall_fig}). Further KB details are in the supplementary material.
}
In Fig.~\ref{fig:pool_size_performance_fig}, performances under three constructed KBs are plotted.
\rev{
Overall, RMCD consistently achieves the best results from the weakest to strongest retrieval results.
Methods using only a single context (RAG, SCD) degrade sharply as retrieval becomes difficult, as they cannot leverage multiple contexts, widening the gap between RMCD and SCD from 6.7\% to 18\% on 1.7K and 2M KBs.
Meanwhile, multi-context methods suffer from irrelevant contexts and underperform RMCD across all KB sizes.
Overall, RMCD proves robust, performing best regardless of retrieval quality through proper reflection and deflection.
}

\input{tab_fig_tex/compare_oracle_tab}
\jh{
\noindent\textbf{Performance with an oracle retriever.}
}
\input{tab_fig_tex/ablation_tables}
\input{tab_fig_tex/qualitative_fig}
\jh{In Tab.~\ref{tab:oracle_retriever_tab}, results with an \q{oracle} context are reported, which denotes the context in the KB corresponding to a question.}
Results show that RMCD achieves the highest accuracy of 38.0, outperforming all other methods, showing that RMCD enables additional gains even when an oracle context is always provided.
To be specific, for RAG and SCD, the oracle context is provided as input.
For methods requiring multiple contexts, the oracle context in $\mathcal{C}$ (if exists), or the least relevant context $c_5$ is firstly removed.
Then, the indices of remaining contexts are shifted right, and the oracle is set as $c_1$. 
We assign $s_{c_1}$ the same retrieval score as $s_{c_2}$, the next-best context score, to keep comparisons fair while larger $s_{c_1}$ is beneficial for RMCD. 
\jh{RMCD outperforming SCD shows the benefit of deflecting irrelevant context, yielding greater gains than merely amplifying the effect of an oracle context.
Also}, the weak performance of methods using multiple contexts suggests that RMCD's success is not solely due to additional contexts. 
The result showcases the necessity of RMCD orthogonal to the development of a better retriever, as it enables additional gains even when the challenging goal of developing a \textit{perfect} retriever is achieved.


\subsection{Ablation study}
\noindent\textbf{Ablation study on the deflection strategy of RMCD.}
To validate the importance of the \q{deflection}, we provide the result of RMCD without deflection in Tab.~\ref{tab:deflection_ablation_tab}.
In detail, logits with negative context weights $\alpha_j$, which are logits to be deflected, are excluded from the calculation of the final logit (Eq.~\ref{eq:mcd_with_constraint}).
As reported, although RMCD without deflection still outperforms SCD by reflecting multiple promising contexts, it underperforms the full RMCD with deflection.
Results validate that explicitly deducting logits with irrelevant contexts, rather than simply ignoring them contributes to better performance by opposing influences of those contexts, demonstrating the importance of deflection strategy.

\noindent\textbf{Ablation study on constraint context set $\mathcal{C}^c$.}
\rev{In Tab.~\ref{tab:constraint_set_ablation_tab}, results with different design choices to build $\mathcal{C}^c$, a constraint context set used for ensembled plausibility constraint, are reported.
Here,
}
\q{best context} denotes only $c_1$ with the highest score consisting $\mathcal{C}^c$, while \q{all contexts} denotes that the whole retrieved context set $\mathcal{C}$ is used as $\mathcal{C}^c$.
As reported, selecting promising contexts based on threshold shows better results than utilizing only a single best context or every context retrieved.
Thus, we conjecture that the proper number of contexts implements more reliable constraints by ensuring the diversity and quality of outputs.
In the supplementary material, we also present ablation results without the plausibility constraint, showing that the resulting performance drop underscores the constraint's importance.

\noindent\textbf{Ablation study on context weight.}
Tab.~\ref{tab:context_weight_ablation_tab} presents the results of different choices to determine the context weight $\alpha_j$.
\q{Uniform} is a case where $\alpha_j$ is set to the value of $j$-th interval uniformly split between $M$ and $m$ not considering the retrieval score $s_{c_j}$.
Absolute score means that the retrieval scores $s_{c_1}$ and $s_{c_j}$ are directly used instead of relative context scores $w_{c_1}$ and $w_{c_j}$ in Eq.~\eqref{eq:multi_cd_coef}.
Relative score denotes the proposed method using relative scores $w_{c_j}$ to determine $\alpha_j$ (Eq.~\eqref{eq:multi_cd_coef}).
The worst performance of uniform weights shows the importance of relevance-aware determination of $\alpha_j$.
Also, adopting a relative score performs better than an absolute score, showing the validity of the design.

\subsection{Qualitative examples}
In Fig.~\ref{fig:qual_fig}, qualitative examples of prediction results along with retrieved \jh{contexts} are illustrated.
As only $c_1$ is provided to RAG, it fails to reflect the evidence in $c_2$ (\dq{look similar to a dandelion}, \dq{subspecies of a common chimpanzee}), thus predicting the wrong answer.
On the other hand, methods using multiple contexts are disturbed by irrelevant information in $c_5$ (\dq{sandal or slipper}, \dq{elephant}), also resulting in wrong predictions.
In contrast, RMCD not only reflects $c_2$ but also deflects $c_4$ and $c_5$, shown by corresponding $\alpha_j$, thereby predicting correct answers by successfully leveraging multiple contexts based on their relevance.

%% file: tab_fig_tex/compare_oracle_tab.tex

\begin{table}[t!]
    \centering
    \renewcommand{\arraystretch}{0.9}
    \setlength{\tabcolsep}{12pt}
    \begin{adjustbox}{max width=\linewidth}
    \begin{tabular}{l|cc|c}
        \toprule
        Method & $n$ & Recall@1 & InfoSeek\\
        \midrule
        Unconditional & 0 & N/A &7.9 \\
        RAG & 1 & 100.0  & 36.3 \\
        SCD & 1 & 100.0  & 37.2 \\
        Consistency & 5 & 100.0 &  22.0 \\
        Max Probability & 5 & 100.0 &  26.8 \\
        Concat & 5 & 100.0 & 22.1 \\
        \midrule
        \rowcolor{lightblue}
        RMCD \textit{(Ours)} & 5 & 100.0 &  \textbf{38.0} \\
        \bottomrule
    \end{tabular}
    \end{adjustbox}
    \vspace{-0.3cm}
    \caption{\textbf{Performance with an oracle context.} $n$: number of contexts used, N/A: not using retrieval results.}
    \label{tab:oracle_retriever_tab}
    \vspace{-0.6cm}
\end{table}

%% file: tab_fig_tex/ablation_tables.tex
\begin{table*}[t!]
\centering
\begin{minipage}{0.33\textwidth}
    \centering
    \small
    \begin{tabular}{l|cc}
        \toprule
        Method & E-VQA & InfoSeek \\
        \midrule
        SCD & 36.0 & 19.6 \\
        RMCD (no defl.) & 38.8 & 20.8 \\
        \midrule
        \rowcolor{lightblue}
        RMCD \textit{(Ours)} & \textbf{39.5} & \textbf{21.0} \\
        \bottomrule
    \end{tabular}
    \vspace{-0.3cm}
    \caption{\textbf{Ablation study on the deflection strategy.} no defl.: RMCD without deflection.}
    \label{tab:deflection_ablation_tab}
\end{minipage}%
\hfill
\begin{minipage}{0.33\textwidth}
    \centering
    \small
    \begin{tabular}{l|cc}
        \toprule
        Method & E-VQA & InfoSeek \\
        \midrule
        Best Context & 36.5 & 20.9 \\
        All Contexts & 37.3 & 20.8 \\
        \midrule
        \rowcolor{lightblue}
        RMCD \textit{(Ours)} & \textbf{39.5} & \textbf{21.0} \\
        \bottomrule
    \end{tabular}
    \vspace{-0.3cm}
    \caption{\textbf{Ablation study on constraint context set $\mathcal{C}^c$ for constraint.}}
    \label{tab:constraint_set_ablation_tab}
\end{minipage}%
\hfill
\begin{minipage}{0.33\textwidth}
    \centering
    \small
    \begin{tabular}{l|cc}
        \toprule
        Method & E-VQA & InfoSeek \\
        \midrule
        Uniform & 35.4 & 20.2 \\
        Absolute & 35.7 & 20.9 \\
        \midrule
        \rowcolor{lightblue}
        Relative \textit{(Ours)} & \textbf{39.5} & \textbf{21.0} \\
        \bottomrule
    \end{tabular}
    \vspace{-0.3cm}
    \caption{\textbf{Ablation study on methods to obtain context weight $\alpha_j$.}}
    \label{tab:context_weight_ablation_tab}
\end{minipage}
\vspace{-0.3cm}
\end{table*}

%% file: tab_fig_tex/qualitative_fig.tex
\begin{figure*}[t!]
\begin{center}
\includegraphics[width=1.0\textwidth]{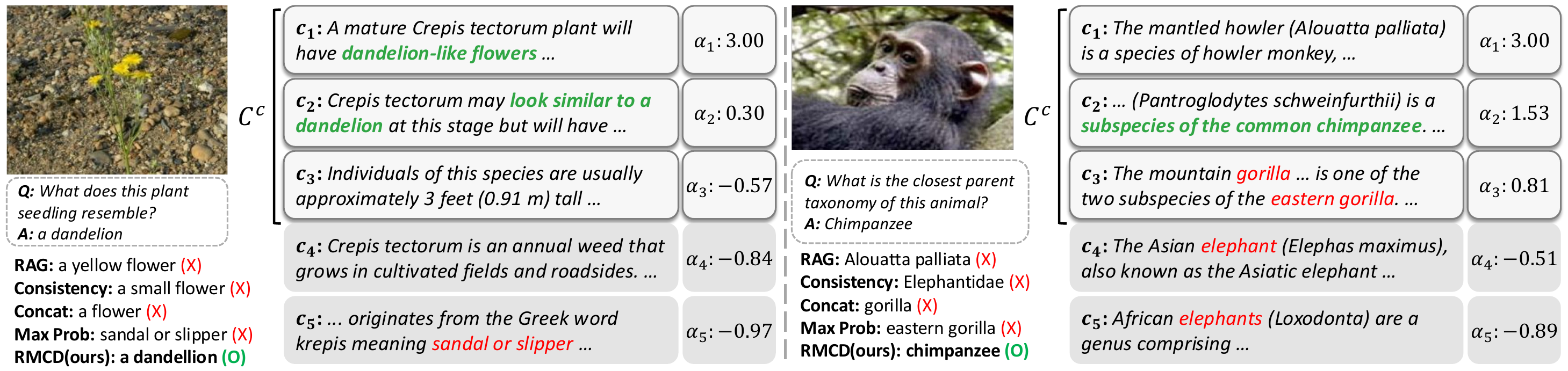}
\end{center}
\vspace{-0.45cm}
\caption{
\textbf{Qualitative Examples.}
Predictions of each method are illustrated along with retrieved contexts $c_j \in \mathcal{C}$, their corresponding context weights $\alpha_j$, and the constraint context set $\mathcal{C}^c$.
{\color{ForestGreen}{Relevant}} and {\color{red}{irrelevant}} information in contexts are marked {\color{ForestGreen}{green}} and {\color{red}{red}}, respectively.
}
\label{fig:qual_fig}
\vspace{-0.4cm}
\end{figure*}

%% file: sec/6_conclusion.tex
\section{Conclusion}
In this paper, we propose Relevance-aware Multi-context Contrastive Decoding (RMCD), a novel decoding method for RAG.
RMCD outputs a final prediction by combining logits obtained with multiple contexts, where each logit is weighted by its relevance to the question.
Thus, it amplifies the effect of relevant contexts and counteracts the effect of irrelevant contexts.
\rev{Applied in a training-free manner, RMCD consistently improves multiple LVLMs on three knowledge-intensive visual question answering benchmarks, achieving the best results.}

%% file: sec/X_suppl.tex
\appendix
\clearpage
\setcounter{page}{1}
\maketitlesupplementary

\section{Ablation study on context score of an empty context}
\input{tab_fig_tex/uncond_ablation_tab}
\input{tab_fig_tex/ablation_without_constraint}
\input{tab_fig_tex/llm_only_tab}
In Tab.~\ref{tab:empty_context_tab}, analysis of choices for a context score $s_{c_{n+1}}$ of an empty context $c_{n+1}$ is provided.
Note that a result with \q{empty context} $c_{n+1}$ denotes a case where no additional context is fed into an LVLM (\ie unconditional generation), and $s_{c_{n+1}}$ denotes the retrieval score corresponding to $c_{n+1}$.
We experimented with three design choices to determine $s_{c_{n+1}}$.
First, \q{Exclude} denotes that an empty context is excluded from calculating the final logit (Eq.8 and Eq.10 of the main paper).
Second, \q{Mean} denotes that $s_{c_{n+1}}$ is set as the mean of $s_{c_j}$ of other contexts.
Finally, \q{$-\infty$} is where $s_{c_{n+1}}$ is set as $-\infty$, which is the choice we adopted in the main experiments.
As reported in the table, we empirically found that setting $s_{c_{n+1}}$ to $-\infty$ worked the best, while other choices have also shown decent performances.


\section{Ablation study on ensembled plausibility constraint}
In Tab.~\ref{tab:ablation_withou_constraint_tab}, ablation results without the ensembled plausibility constraint are reported.
The ensembled plausibility constraint elaborated through Eqs.~9 to 13 of the main paper is an additional constraint that complements RMCD by restricting an LVLM from sampling highly unlikely tokens, thereby ensuring better generation results.
As reported, RMCD without the proposed plausibility constraint shows sub-optimal performances, showing the importance of implementing the proposed constraint.
Note that Contrastive Decoding-based methods performing worse without proper plausibility constraints is a common phenomenon, where the same result is also reported in the original CD paper~\cite{li2022contrastive}.

\section{Extending RMCD to text-based QA tasks}
In Tab.~\ref{tab:tab_nq_ic_ralm}, we report the performance of RMCD on Natural Questions~\cite{kwiatkowski2019natural} benchmark, an open-domain question-answering benchmark purely based on natural languages.
In detail, we apply RMCD to IC-RALM~\cite{ram2023context}, an RAG pipeline based on Large Language Models (LLMs).
The results show that replacing the decoding method of IC-RALM with RMCD results in an additional gain of 2.5 points in accuracy, showing that RMCD can be also extended to pure NLP tasks to improve existing RAG pipelines.




\section{Construction of KBs for InfoSeek}
Since the Wikipedia Knowledge Base (KB) used as a retrieval source in the original InfoSeek~\cite{chen2023can} paper is not provided by the authors, we construct the KB using the Wikipedia dump that the authors have provided.
Similar to the original paper, we first filter out entities in the Wikipedia dump without images, resulting in a filtered dump with a size of 2M.
Then, we first construct the smallest KB by selecting 1.7K entities related to samples in InfoSeek validation set from the filtered dump based on annotations.
Note that the 1.7K KB is a default for every experiment in the paper, except for experiments in Fig.~4.
We additionally construct 100K and 2M KBs to test the robustness of decoding methods against retrieval quality.
After constructing the 1.7K KB, we randomly sample non-relevant entities from the remaining entities, which are added to the 1.7K KB forming a KB with a size of 100K.
Also, we simply define the largest 2M KB by using every entity in the filtered dump.
As the ratio of entities irrelevant to QA pairs increases in larger KB, retrieval gets more difficult, as shown in the decline of the retrieval Recall@$k$ metric in Fig.~3 of the main paper.

\section{Retrieval procedure}
\input{tab_fig_tex/sensitivity_tabs_1}
\input{tab_fig_tex/sensitivity_tabs_2}
\noindent\textbf{InfoSeek.} For InfoSeek~\cite{chen2023can}, we use the image corresponding to a QA pair as a query.
A query feature is obtained with the image encoder of CLIP-ViT-B/32~\cite{radford2021learning}.
Then, features of every document in the KB are defined as a text feature of a summary corresponding to a document, obtained with a CLIP text encoder.
Based on the obtained query and document features, the retrieval score between a query and a document is set as an inner product between the query feature and the document feature.
Finally, we retrieve $n$ documents with the highest retrieval scores, taking their summaries as retrieved contexts, where the retrieval scores $s_{c_j}$ between a sample and a context $c_j$ are set as retrieval scores used. 

\noindent\textbf{Encyclopedic-VQA.} For Encyclopedic-VQA~\cite{mensink2023encyclopedic}, we define the retrieval target documents using retrieval results with Google Lens~\cite{googlelens} provided by authors, since CLIP-based retrieval is shown to be very challenging~\cite{mensink2023encyclopedic} on Encyclopedic-VQA.
In detail, we adopt five documents with the highest Lens retrieval score as retrieval target documents.
In the case where less than five Lens retrieval results exist, we additionally retrieve target documents with CLIP scores identical to the same process in InfoSeek until obtaining five target documents without duplicates.
Documents retrieved with CLIP score are appended to Lens retrieval results.
Since the document summary does not exist in KB for Encyclopedic-VQA, an additional step of determining passages in documents (\ie paragraph) to retrieve is required.
Among top-$k$ documents having at least $n$ passages in total, a retrieval score between a sample and a passage is defined with a BM25 similarity~\cite{robertson2009probabilistic} between a question and the passage.
Based on calculated BM25 similarities, we retrieve $n$ passages with the highest similarity, where the retrieval score $s_{c_j}$ between a sample and a context $c_j$ is set as a BM25 similarity.

\noindent\textbf{OK-VQA.} For OK-VQA, we directly adopt the retrieval results used in the baselines~\cite{lin2022retrieval,lin2024fine} where RMCD is implemented, without modification.
Those retrieval results are obtained from Google Search Corpus~\cite{luo2021weakly} with retrievers proposed in each work.

\rev{\section{Analysis on hyperparameters}
\noindent\textbf{Sensitivity analysis.}}
From Tab.~\ref{tab:sensitivity_M_tab} to~\ref{tab:sensitivity_tau_2_tab}, we report performances on Encyclopedic-VQA under different hyperparameters.
Overall, RMCD shows robust performance regardless of the choice of hyperparameter.
Concretely, it consistently outperforms the second-best method \q{concat}, referred to as \q{Non-RMCD Best} in tables, regardless of the hyperparameters.
In terms of $M$ and $m$ controlling the maximum and minimum value of context weights, we found that larger values show better performance in general.
Regarding $\tau_1$ and $\tau_2$, which control softmax temperature, results show that adopting $\tau_1 = 1$ or $\tau_2 = 1$ still outperforms the second-best performance.
Such a result shows that adopting $\tau_1$ or $\tau_2$ is an optional choice for better performance, as setting a softmax temperature to $1$ is equivalent to eliminating those hyperparameters.
Finally, we observe that the choice of $\gamma$ and $\beta$ have minimal impact on the final performance.

\rev{
\noindent\textbf{Selection guideliens.}
Here, we provide guidelines for hyperparameter selection, although RMCD is generally robust to hyperparameters.
For reflection strength $M$, smaller values are preferable under poor retrieval, thus we adopt relatively low $M=3.0$ for OK-VQA with noisy web corpus.
Setting $m$ to overly small values (below $-1.5$) leads to suboptimal performance because it penalizes relevant contexts.
For $\gamma$ and $\beta$, performance remains stable across the ranges reported in Tab.\ref{tab:sensitivity_gamma_tab} and Tab.\ref{tab:sensitivity_beta_tab}.
Softmax temperatures above $1.0$ work well for $\gamma_1$, while values below $1.0$ are better for $\gamma_2$.
}

\section{Statistics of retrieval results}
\input{tab_fig_tex/example_prompt_tab}

\input{tab_fig_tex/wiki_kb_split_stats}

\input{tab_fig_tex/encyclopedic_recall_tab}

\noindent\textbf{InfoSeek.} In Tab.~\ref{tab:wikipedia_kb_split_stats_tab}, retrieval results on InfoSeek validation set are reported.
In detail, Recall@$k$, where $k = 1, 3, 5$ are reported in document-level.
Since we construct the KB with three different sizes, results are reported on every KB.

\paragraph{Encyclopedic-VQA.}
We report the passage-level and document-level retrieval results on Encyclopedic-VQA in Recall@$k$ (R@$k$), where $k = 1, 3, 5$.
To be specific, only the retrieval results on the one-hop split are reported since two-hop questions require two different documents to answer the question, making it challenging to define the accurate Recall@$k$ metric.
Results are reported in Tab.~\ref{tab:encyclopedic_recall_tab}.

\section{Further implementation details}
\input{tab_fig_tex/supp_qualitative_fig}
\noindent\textbf{Encyclopedic-VQA preprocessing.}
We find that image URLs corresponding to some samples in the test set of Encyclopedic-VQA~\cite{mensink2023encyclopedic} are not accessible due to their expiration.
Therefore, we excluded 38 samples from the test set whose image was not accessible.
Every experimental result in this paper is reported with such samples excluded.

\noindent\textbf{Details about baseline LVLMs.}
As baseline LVLMs for InfoSeek and Encyclopedic-VQA benchmarks, we adopt InternVL-2.5~\cite{chen2024expanding}, BLIP-2~\cite{li2023blip} and LLaVA-1.5~\cite{liu2023improved}.
We use the official pre-trained weights and implementations of InternVL-2.5~\cite{chen2024expanding} by the by the Hugging Face transformers library~\cite{wolf2020transformers}, BLIP-2 by the LAVIS library~\cite{li2022lavis}, and LLaVA-1.5 by the Hugging Face transformers library~\cite{wolf2020transformers}.
As an image encoder of BLIP-2, ViT-g/14 variant of EVA-CLIP~\cite{fang2023eva} is adopted.
As a base Language Model (LM), OPT-6.7B~\cite{zhang2022opt}, T5-XL, and T5-XXL~\cite{chung2024scaling} are used for BLIP-2.
The input image size for BLIP-2 is set to $224 \times 224$ pixels, and the model is loaded in BF16 precision.
As an image encoder of LLaVA-1.5, CLIP-ViT-L-336px~\cite{radford2021learning} is adopted.
As a base LM, Vicuna-13B~\cite{vicuna2023} is utilized.
The input image size for LLaVA-1.5 is set to $336 \times 336$ pixels.
The model is loaded in BF16 precision, except for the Vicuna-13B language model, which uses 4-bit quantization~\cite{dettmers2022gpt3}.
As baseline LVLMs for OK-VQA, we use the pre-trained models provided by the authors~\cite{lin2022retrieval,lin2024fine}, where the model architectures are based on T5-XL and BLIP-2 with T5-XL as base LM, respectively.
Nucleus sampling~\cite{holtzman2019curious} with $p=0.9$ and a temperature of 1.0 is applied for sampling.

\noindent\textbf{Input prompt.}
In Tab.~\ref{tab:prompt_template_examples_tab}, prompt templates for unconditional decoding and retrieval-based methods (\ie RAG, SCD, consistency, concat, max probability, and RMCD) are provided.
For the unconditional generation, we adopt default prompts for question-answering tasks from the original implementations of BLIP-2~\cite{li2023blip} and LLaVA-1.5~\cite{liu2023improved}.
For retrieval-based methods, we simply append the retrieved context after the question.
As an input prompt of RA-VQA~\cite{lin2022retrieval} and FLMR~\cite{lin2024fine} for the OK-VQA benchmark, we directly adopt the baselines' prompts for retrieval-based generation.

\noindent\textbf{Evaluation metrics.}
In \textbf{InfoSeek}, the total accuracy is defined as the harmonic mean of accuracy on two validation splits, \textsc{unseen question} and \textsc{unseen entity}.
\textsc{Unseen question} consists of the QA pairs that do not exist in InfoSeek train set.
\textsc{Unseen entity} consists of QA pairs where the entity corresponding to the QA pair does not exist in the InfoSeek train set.
Since RMCD is a training-free method, the two divisions are not significantly important.
Still, we report performance on both splits for reference.
Accuracy on each split is the average accuracy of three question types: \textsc{string}, \textsc{numerical}, and \textsc{time}.
For \textsc{string} and \textsc{time} question types which contain multiple possible answers, evaluation is done following conventional VQA practices~\cite{goyal2017making}.
For \textsc{numerical} questions asking for detailed numbers, the Related Accuracy from~\cite{masry2022chartqa} is utilized, which allows an error within a $10\%$ tolerance range.

In \textbf{Encyclopedic-VQA}~\cite{mensink2023encyclopedic}, questions are classified into three types: one-hop, multi-answer, and two-hop.
While the original paper mainly reported performance on one-hop questions in the majority of experiments, we additionally reported the performance on every question as a reference.
Each prediction in Encyclopedic-VQA except questions of the \q{multi-answer} type is evaluated with BERT Matching (BEM)~\cite{bulian2022tomayto}, which employs BERT~\cite{devlin2019bert} to classify whether the prediction is correct based on the given answer.
In detail, if the BEM score between a prediction and the answer is over 0.5, the prediction is considered correct.
For questions of the \q{multi-answer} type, model predictions are converted into a set of strings, and the intersection-over-union (IoU) between the prediction set and answer set is calculated.
If the IoU is over 0.5, the prediction is considered correct.
Otherwise, the BEM score between concatenated predictions and concatenated answers is used to evaluate predictions.

In \textbf{OK-VQA}~\cite{marino2019ok}, we directly follow the evaluation protocol of baselines~\cite{lin2022retrieval, lin2024fine} where the VQA score~\cite{marino2019ok} is adopted as an accuracy metric.

\section{Further qualitative examples}
We provide further qualitative examples in Fig.~\ref{fig:supp_qual_fig}.
Examples demonstrate that RMCD effectively reflects relevant contexts while also deflecting irrelevant contexts, therefore generating better results compared to other decoding methods.
For instance, RMCD successfully reflects evidence in relevant contexts (\dq{Rocks from Moon \& Mars}, \dq{located on both banks}) which other decoding methods fail.
Also, unlike other methods affected by wrong evidence existent in multiple irrelevant contexts thereby generating wrong answers (\dq{apophyllite}, \dq{footbridge}), RMCD successfully deflects irrelevant contexts and thus opposes their effects.




%% file: tab_fig_tex/uncond_ablation_tab.tex
\begin{table}[t!]
\centering
    \centering
    \begin{adjustbox}{width=0.6\linewidth}
    \begin{tabular}{l|ccc}
        \toprule
        Method & E-VQA & InfoSeek \\
        \midrule
        Exclude & 36.3 & 20.2 \\
        Mean & 37.2 & 20.2 \\
        \midrule
        \rowcolor{lightblue}
        $-\infty$ \textit{(Ours)} & \textbf{39.5} & \textbf{21.0} \\ 
        \bottomrule
    \end{tabular}
    \end{adjustbox}
    \caption{\textbf{Analysis on choices for $s_{c_{n+1}}$ of an empty context.}}
    \label{tab:empty_context_tab}
\end{table}

%% file: tab_fig_tex/ablation_without_constraint.tex
\begin{table}[t!]
\centering
    \centering
    \begin{adjustbox}{width=0.8\linewidth}
    \begin{tabular}{l|cc}
        \toprule
        Method & E-VQA & InfoSeek \\
        \midrule
        RMCD (No Constraint) & 37.2 & 12.5 \\
        \rowcolor{lightblue}
        RMCD \textit{(Ours)} & \textbf{39.5} & \textbf{21.0} \\
        \bottomrule
    \end{tabular}
    \end{adjustbox}
    \caption{\textbf{Ablation study on ensembled plausibility constraint.}}
    \label{tab:ablation_withou_constraint_tab}
\end{table}

%% file: tab_fig_tex/llm_only_tab.tex
\begin{table}[t!]
\centering
    \centering
    \begin{adjustbox}{width=0.8\linewidth}
    \begin{tabular}{l|cc}
        \toprule
        Method & $n$ & Exact Match \\
        \midrule
        IC-RALM~\cite{ram2023context} (LLaMA-7B) & 5 & 24.1 \\ 
        \rowcolor{lightblue}
        + RMCD \textit{(Ours)} & 5 & \textbf{26.6} \\
        \bottomrule
    \end{tabular}
    \end{adjustbox}
    \caption{\textbf{Results on Natural Questions~\cite{kwiatkowski2019natural} benchmark.}}
    \label{tab:tab_nq_ic_ralm}
\end{table}

%% file: tab_fig_tex/sensitivity_tabs_1.tex
\begin{table*}[t!]
\centering
\begin{minipage}{0.32\linewidth}
    \input{tab_fig_tex/sensitivity_M_tab}
\end{minipage}
\hfill
\begin{minipage}{0.32\linewidth}
    \input{tab_fig_tex/sensitivity_m_tab}
\end{minipage}
\hfill
\begin{minipage}{0.32\linewidth}
    \input{tab_fig_tex/sensitivity_gamma_tab}
\end{minipage}
\end{table*}

%% file: tab_fig_tex/sensitivity_M_tab.tex
\centering
\begin{tabular}{c|c}
\toprule
$M$ & \small Encyclopedic-VQA \\
\midrule
\small Non-RMCD Best & 37.4 \\
\midrule
2.0 & 38.6 \\
3.0 & 39.3 \\
\rowcolor{lightblue}
4.0 & 39.5 \\
5.0 & 39.4 \\
\bottomrule
\end{tabular}
\caption{\textbf{Sensitivity Analysis on $M$.}}
\label{tab:sensitivity_M_tab}

%% file: tab_fig_tex/sensitivity_m_tab.tex
\centering
\begin{tabular}{c|c}
\toprule
$m$ & \small Encyclopedic-VQA \\
\midrule
\small Non-RMCD Best & 37.4 \\
\midrule
-0.5 & 39.4 \\
\rowcolor{lightblue}
-1.0 & 39.5 \\
-1.5 & 39.1 \\
-2.0 & 38.9 \\
\bottomrule
\end{tabular}
\caption{\textbf{Sensitivity Analysis on $m$.}}
\label{tab:sensitivity_m_tab}

%% file: tab_fig_tex/sensitivity_gamma_tab.tex
\centering
\begin{tabular}{c|c}
\toprule
$\gamma$ & \small Encyclopedic-VQA \\
\midrule
\small Non-RMCD Best & 37.4 \\
\midrule
0.05 & 39.3 \\
0.1 & 39.2 \\
\rowcolor{lightblue}
0.3 & 39.5 \\
0.5 & 39.0 \\
\bottomrule
\end{tabular}
\caption{\textbf{Sensitivity Analysis on $\gamma$.}}
\label{tab:sensitivity_gamma_tab}

%% file: tab_fig_tex/sensitivity_tabs_2.tex
\begin{table*}[t!]
\centering
\begin{minipage}{0.32\linewidth}
    \input{tab_fig_tex/sensitivity_beta_tab}
\end{minipage}
\hfill
\begin{minipage}{0.32\linewidth}
    \input{tab_fig_tex/sensitivity_tau_1_tab}
\end{minipage}
\hfill
\begin{minipage}{0.32\linewidth}
    \input{tab_fig_tex/sensitivity_tau_2_tab}
\end{minipage}
\end{table*}

%% file: tab_fig_tex/sensitivity_beta_tab.tex
\centering
\begin{tabular}{c|c}
\toprule
$\beta$ & \small Encyclopedic-VQA \\
\midrule
\small Non-RMCD Best & 37.4 \\
\midrule
0.1 & 39.4 \\
0.15 & 39.3 \\
\rowcolor{lightblue}
0.2 & 39.5 \\
0.3 & 39.1 \\
\bottomrule
\end{tabular}
\caption{\textbf{Sensitivity Analysis on $\beta$.}}
\label{tab:sensitivity_beta_tab}

%% file: tab_fig_tex/sensitivity_tau_1_tab.tex
\centering
\begin{tabular}{c|c}
\toprule
$\tau_1$ & \small Encyclopedic-VQA \\
\midrule
\small Non-RMCD Best & 37.4 \\
\midrule
1.0 & 37.7 \\
2.0 & 39.0 \\
\rowcolor{lightblue}
3.0 & 39.5 \\
4.0 & 39.3
\\
\bottomrule
\end{tabular}
\caption{\textbf{Sensitivity Analysis on $\tau_1$.}}
\label{tab:sensitivity_tau_1_tab}

%% file: tab_fig_tex/sensitivity_tau_2_tab.tex
\centering
\begin{tabular}{c|c}
\toprule
$\tau_2$ & \small Encyclopedic-VQA \\
\midrule
\small Non-RMCD Best & 37.4 \\
\midrule
0.1 & 39.4 \\
\rowcolor{lightblue}
0.5 & 39.5 \\
1.0 & 38.8 \\
2.0 & 38.7 \\
\bottomrule
\end{tabular}
\caption{\textbf{Sensitivity Analysis on $\tau_2$.}}
\label{tab:sensitivity_tau_2_tab}

%% file: tab_fig_tex/example_prompt_tab.tex
\begin{table*}[ht!]
\centering
\begin{adjustbox}{width=\linewidth}
\begin{tabular}{c|c|c}
\toprule
Model                        & Method          & Template\\
\midrule
\multirow{2}{*}[-0.09cm]{BLIP-2~\cite{li2023blip}} & Unconditional   & \small \dq{Question: $<$question$>$, Short answer:}\\
\cmidrule{2-3}
                             & Retrieval-based & \small \dq{Question: $<$question$>$, Context: $<$context$>$ Short answer:}\\
\midrule
\multirow{3}{*}[-0.06cm]{LLaVA-1.5~\cite{liu2024visual}}   & Unconditional   & \small \makecell{\dq{$<$question$>$ Answer the question using a single word or phrase.}}\\
\cmidrule{2-3}
                             & Retrieval-based & \small \makecell{\dq{$<$question$>$ Answer the question using a single word or phrase.\\Context: $<$context$>$}}\\
     \midrule
RA-VQA~\cite{lin2022retrieval} & Retrieval-based & \small \dq{$<$question$>$ $<$caption$>$ $<$objects$>$ $<$context$>$}\\
\midrule
FLMR~\cite{lin2024fine} & Retrieval-based & \small \dq{Question: $<$question$>$ Caption: $<$caption$>$ Object: $<$objects$>$ Knowledge: $<$context$>$ Answer:}\\
\bottomrule
\end{tabular}
\end{adjustbox}
\caption{\textbf{Prompt templates.} $<$question$>$ and $<$context$>$ in templates are replaced with actual questions and contexts, respectively. $<$caption$>$ and $<$object$>$ are replaced with image captioning results and detected objects in~\cite{lin2022retrieval, lin2024fine}.}
\label{tab:prompt_template_examples_tab}
\end{table*}

%% file: tab_fig_tex/wiki_kb_split_stats.tex
\begin{table}[t!]
\centering
\begin{tabular}{c|ccc}
\toprule
KB Size & Recall@1 & Recall@3 & Recall@5 \\
\midrule
1.7K & 45.3 & 63.5 & 70.5 \\ 
100K & 19.5 & 30.6 & 36.1 \\ 
2M & 6.1 & 11.7 & 14.8 \\ 
\bottomrule
\end{tabular}
\caption{\textbf{Retrieval results on InfoSeek validation set.}}
\label{tab:wikipedia_kb_split_stats_tab}
\end{table}

%% file: tab_fig_tex/encyclopedic_recall_tab.tex
\begin{table}[t!]
\centering
\begin{tabular}{c|ccc}
\toprule
Retrieval Target           & Recall@1 & Recall@3 & Recall@5 \\
\midrule
                          Document   & 50.2     & 61.8     & 65.6     \\
                          Passage      & 26.6     & 39.8     & 45.8     \\
\bottomrule
\end{tabular}
\caption{\textbf{Document-level and passage-level retrieval results on Encyclopedic-VQA test set.}}
\label{tab:encyclopedic_recall_tab}
\end{table}

%% file: tab_fig_tex/supp_qualitative_fig.tex
\begin{figure*}[t!]
\begin{center}
\includegraphics[width=1.0\textwidth]{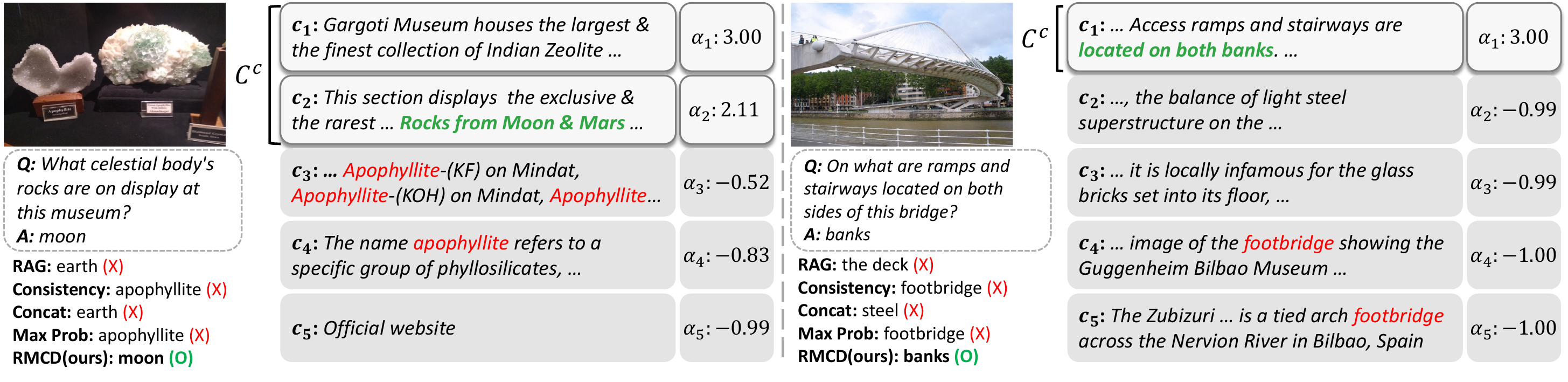}
\end{center}
\caption{
\textbf{Further Qualitative Examples.}
Predictions of each method are illustrated along with retrieved contexts $c_j \in \mathcal{C}$, their corresponding context weights $\alpha_j$, and the constraint context set $\mathcal{C}^c$.
{\color{ForestGreen}{Relevant}} and {\color{red}{irrelevant}} information in contexts are marked {\color{ForestGreen}{green}} and {\color{red}{red}}, respectively.
}
\label{fig:supp_qual_fig}
\end{figure*}

%% file: main.bib
@String(CVPR= {IEEE Conf. Comput. Vis. Pattern Recog.})

@String(ICCV= {Int. Conf. Comput. Vis.})

@String(ECCV= {Eur. Conf. Comput. Vis.})

@String(ICLR = {Int. Conf. Learn. Represent.})

@String(AAAI = {AAAI})

@String(CVPR  = {CVPR})

@String(ICCV  = {ICCV})

@String(ECCV  = {ECCV})

@String(ICLR  = {ICLR})

@inproceedings{li2022contrastive,
  title={Contrastive decoding: Open-ended text generation as optimization},
  author={Li, Xiang Lisa and Holtzman, Ari and Fried, Daniel and Liang, Percy and Eisner, Jason and Hashimoto, Tatsunori and Zettlemoyer, Luke and Lewis, Mike},
  booktitle={ACL},
  year={2023}
}

@inproceedings{chuang2024dola,
  title={Dola: Decoding by contrasting layers improves factuality in large language models},
  author={Chuang, Yung-Sung and Xie, Yujia and Luo, Hongyin and Kim, Yoon and Glass, James and He, Pengcheng},
  booktitle={ICLR},
  year={2024}
}

@inproceedings{leng2023mitigating,
  title={Mitigating object hallucinations in large vision-language models through visual contrastive decoding},
  author={Leng, Sicong and Zhang, Hang and Chen, Guanzheng and Li, Xin and Lu, Shijian and Miao, Chunyan and Bing, Lidong},
  booktitle={CVPR},
  year={2024}
}

@inproceedings{kiminstructive,
  title={Instructive Decoding: Instruction-Tuned Large Language Models are Self-Refiner from Noisy Instructions},
  author={Kim, Taehyeon and Kim, Joonkee and Lee, Gihun and Yun, Se-Young},
  booktitle={ICLR},
  year={2024}
}

@inproceedings{li2023blip,
  title={Blip-2: Bootstrapping language-image pre-training with frozen image encoders and large language models},
  author={Li, Junnan and Li, Dongxu and Savarese, Silvio and Hoi, Steven},
  booktitle={ICML},
  year={2023},
}

@inproceedings{liu2024visual,
  title={Visual instruction tuning},
  author={Liu, Haotian and Li, Chunyuan and Wu, Qingyang and Lee, Yong Jae},
  booktitle={NeurIPS},
  year={2023}
}

@inproceedings{dai2024instructblip,
  title={Instructblip: Towards general-purpose vision-language models with instruction tuning},
  author={Dai, Wenliang and Li, Junnan and Li, Dongxu and Tiong, Anthony Meng Huat and Zhao, Junqi and Wang, Weisheng and Li, Boyang and Fung, Pascale N and Hoi, Steven},
  booktitle={NeurIPS},
  year={2023}
}

@article{ye2023mplug,
  title={mplug-owl: Modularization empowers large language models with multimodality},
  author={Ye, Qinghao and Xu, Haiyang and Xu, Guohai and Ye, Jiabo and Yan, Ming and Zhou, Yiyang and Wang, Junyang and Hu, Anwen and Shi, Pengcheng and Shi, Yaya and others},
  journal={arXiv preprint arXiv:2304.14178},
  year={2023}
}

@inproceedings{luo2021weakly,
  title={Weakly-supervised visual-retriever-reader for knowledge-based question answering},
  author={Luo, Man and Zeng, Yankai and Banerjee, Pratyay and Baral, Chitta},
  booktitle={EMNLP},
  year={2021}
}

@inproceedings{liu2023improved,
  title={Improved baselines with visual instruction tuning},
  author={Liu, Haotian and Li, Chunyuan and Li, Yuheng and Lee, Yong Jae},
  booktitle={CVPR},
  year={2024}
}

@article{zhang2022opt,
  title={Opt: Open pre-trained transformer language models},
  author={Zhang, Susan and Roller, Stephen and Goyal, Naman and Artetxe, Mikel and Chen, Moya and Chen, Shuohui and Dewan, Christopher and Diab, Mona and Li, Xian and Lin, Xi Victoria and others},
  journal={arXiv preprint arXiv:2205.01068},
  year={2022}
}

@article{chung2024scaling,
  title={Scaling instruction-finetuned language models},
  author={Chung, Hyung Won and Hou, Le and Longpre, Shayne and Zoph, Barret and Tay, Yi and Fedus, William and Li, Yunxuan and Wang, Xuezhi and Dehghani, Mostafa and Brahma, Siddhartha and others},
  journal={JMLR},
  year={2024}
}

@inproceedings{holtzman2019curious,
  title={The curious case of neural text degeneration},
  author={Holtzman, Ari and Buys, Jan and Du, Li and Forbes, Maxwell and Choi, Yejin},
  booktitle={ICLR},
  year={2020}
}

@inproceedings{chen2023can,
  title={Can pre-trained vision and language models answer visual information-seeking questions?},
  author={Chen, Yang and Hu, Hexiang and Luan, Yi and Sun, Haitian and Changpinyo, Soravit and Ritter, Alan and Chang, Ming-Wei},
  booktitle={EMNLP},
  year={2023}
}

@inproceedings{mensink2023encyclopedic,
  title={Encyclopedic VQA: Visual questions about detailed properties of fine-grained categories},
  author={Mensink, Thomas and Uijlings, Jasper and Castrejon, Lluis and Goel, Arushi and Cadar, Felipe and Zhou, Howard and Sha, Fei and Araujo, Andr{\'e} and Ferrari, Vittorio},
  booktitle={ICCV},
  year={2023}
}

@inproceedings{lewis2020retrieval,
  title={Retrieval-augmented generation for knowledge-intensive nlp tasks},
  author={Lewis, Patrick and Perez, Ethan and Piktus, Aleksandra and Petroni, Fabio and Karpukhin, Vladimir and Goyal, Naman and K{\"u}ttler, Heinrich and Lewis, Mike and Yih, Wen-tau and Rockt{\"a}schel, Tim and others},
  booktitle={NeurIPS},
  year={2020}
}

@inproceedings{marino2019ok,
  title={Ok-vqa: A visual question answering benchmark requiring external knowledge},
  author={Marino, Kenneth and Rastegari, Mohammad and Farhadi, Ali and Mottaghi, Roozbeh},
  booktitle={CVPR},
  year={2019}
}

@inproceedings{radford2021learning,
  title={Learning transferable visual models from natural language supervision},
  author={Radford, Alec and Kim, Jong Wook and Hallacy, Chris and Ramesh, Aditya and Goh, Gabriel and Agarwal, Sandhini and Sastry, Girish and Askell, Amanda and Mishkin, Pamela and Clark, Jack and others},
  booktitle={ICML},
  year={2021},
}

@article{robertson2009probabilistic,
  title={The probabilistic relevance framework: BM25 and beyond},
  author={Robertson, Stephen and Zaragoza, Hugo and others},
  journal={Foundations and Trends{\textregistered} in Information Retrieval},
  year={2009},
}

@inproceedings{masry2022chartqa,
  title={Chartqa: A benchmark for question answering about charts with visual and logical reasoning},
  author={Masry, Ahmed and Long, Do Xuan and Tan, Jia Qing and Joty, Shafiq and Hoque, Enamul},
  booktitle={ACL Findings},
  year={2022}
}

@inproceedings{bulian2022tomayto,
  title={Tomayto, tomahto. beyond token-level answer equivalence for question answering evaluation},
  author={Bulian, Jannis and Buck, Christian and Gajewski, Wojciech and Boerschinger, Benjamin and Schuster, Tal},
  booktitle={EMNLP},
  year={2022}
}

@article{o2023contrastive,
  title={Contrastive decoding improves reasoning in large language models},
  author={O'Brien, Sean and Lewis, Mike},
  journal={arXiv preprint arXiv:2309.09117},
  year={2023}
}

@inproceedings{devlin2019bert,
  title={BERT: Pre-training of Deep Bidirectional Transformers for Language Understanding},
  author={Devlin, Jacob and Chang, Ming-Wei and Lee, Kenton and Toutanova, Kristina},
  booktitle={NAACL-HLT},
  year={2019}
}

@inproceedings{zhu2023minigpt,
  title={Minigpt-4: Enhancing vision-language understanding with advanced large language models},
  author={Zhu, Deyao and Chen, Jun and Shen, Xiaoqian and Li, Xiang and Elhoseiny, Mohamed},
  booktitle={ICLR},
  year={2024}
}

@inproceedings{cha2023honeybee,
  title={Honeybee: Locality-enhanced projector for multimodal llm},
  author={Cha, Junbum and Kang, Wooyoung and Mun, Jonghwan and Roh, Byungseok},
  booktitle={CVPR},
  year={2024}
}

@inproceedings{zhang2023prompt,
  title={Prompt Highlighter: Interactive Control for Multi-Modal LLMs},
  author={Zhang, Yuechen and Qian, Shengju and Peng, Bohao and Liu, Shu and Jia, Jiaya},
  booktitle={CVPR},
  year={2024}
}

@inproceedings{wang2022self,
  title={Self-consistency improves chain of thought reasoning in language models},
  author={Wang, Xuezhi and Wei, Jason and Schuurmans, Dale and Le, Quoc and Chi, Ed and Narang, Sharan and Chowdhery, Aakanksha and Zhou, Denny},
  booktitle={ICLR},
  year={2023}
}

@inproceedings{lin2024fine,
  title={Fine-grained late-interaction multi-modal retrieval for retrieval augmented visual question answering},
  author={Lin, Weizhe and Chen, Jinghong and Mei, Jingbiao and Coca, Alexandru and Byrne, Bill},
  booktitle={NeurIPS},
  year={2023}
}

@misc{googlelens,
  title = {Google lens},
  howpublished = {Web interface available at \url{https://images.google.com}},
}

@inproceedings{goyal2017making,
  title={Making the v in vqa matter: Elevating the role of image understanding in visual question answering},
  author={Goyal, Yash and Khot, Tejas and Summers-Stay, Douglas and Batra, Dhruv and Parikh, Devi},
  booktitle={CVPR},
  year={2017}
}

@article{li2022lavis,
  title={Lavis: A library for language-vision intelligence},
  author={Li, Dongxu and Li, Junnan and Le, Hung and Wang, Guangsen and Savarese, Silvio and Hoi, Steven CH},
  journal={arXiv preprint arXiv:2209.09019},
  year={2022}
}

@inproceedings{fang2023eva,
  title={Eva: Exploring the limits of masked visual representation learning at scale},
  author={Fang, Yuxin and Wang, Wen and Xie, Binhui and Sun, Quan and Wu, Ledell and Wang, Xinggang and Huang, Tiejun and Wang, Xinlong and Cao, Yue},
  booktitle={CVPR},
  year={2023}
}

@inproceedings{dettmers2022gpt3,
  title={Gpt3. int8 (): 8-bit matrix multiplication for transformers at scale},
  author={Dettmers, Tim and Lewis, Mike and Belkada, Younes and Zettlemoyer, Luke},
  booktitle={NeurIPS},
  year={2022}
}

@inproceedings{hu2023open,
  title={Open-domain visual entity recognition: Towards recognizing millions of wikipedia entities},
  author={Hu, Hexiang and Luan, Yi and Chen, Yang and Khandelwal, Urvashi and Joshi, Mandar and Lee, Kenton and Toutanova, Kristina and Chang, Ming-Wei},
  booktitle={ICCV},
  year={2023}
}

@inproceedings{wolf2020transformers,
  title={Transformers: State-of-the-art natural language processing},
  author={Wolf, Thomas and Debut, Lysandre and Sanh, Victor and Chaumond, Julien and Delangue, Clement and Moi, Anthony and Cistac, Pierric and Rault, Tim and Louf, R{\'e}mi and Funtowicz, Morgan and others},
  booktitle={EMNLP: System Demonstrations},
  year={2020}
}

@inproceedings{shah2019kvqa,
  title={Kvqa: Knowledge-aware visual question answering},
  author={Shah, Sanket and Mishra, Anand and Yadati, Naganand and Talukdar, Partha Pratim},
  booktitle={AAAI},
  year={2019}
}

@inproceedings{jain2021select,
  title={Select, substitute, search: A new benchmark for knowledge-augmented visual question answering},
  author={Jain, Aman and Kothyari, Mayank and Kumar, Vishwajeet and Jyothi, Preethi and Ramakrishnan, Ganesh and Chakrabarti, Soumen},
  booktitle={SIGIR},
  year={2021}
}

@inproceedings{hu2023reveal,
  title={Reveal: Retrieval-augmented visual-language pre-training with multi-source multimodal knowledge memory},
  author={Hu, Ziniu and Iscen, Ahmet and Sun, Chen and Wang, Zirui and Chang, Kai-Wei and Sun, Yizhou and Schmid, Cordelia and Ross, David A and Fathi, Alireza},
  booktitle={CVPR},
  year={2023}
}

@inproceedings{gui2021kat,
  title={Kat: A knowledge augmented transformer for vision-and-language},
  author={Gui, Liangke and Wang, Borui and Huang, Qiuyuan and Hauptmann, Alex and Bisk, Yonatan and Gao, Jianfeng},
  booktitle={NACCL-HLT},
  year={2022}
}

@inproceedings{asai2023self,
  title={Self-rag: Learning to retrieve, generate, and critique through self-reflection},
  author={Asai, Akari and Wu, Zeqiu and Wang, Yizhong and Sil, Avirup and Hajishirzi, Hannaneh},
  booktitle={ICLR},
  year={2024}
}

@inproceedings{shi2023large,
  title={Large language models can be easily distracted by irrelevant context},
  author={Shi, Freda and Chen, Xinyun and Misra, Kanishka and Scales, Nathan and Dohan, David and Chi, Ed H and Sch{\"a}rli, Nathanael and Zhou, Denny},
  booktitle={ICML},
  year={2023},
}

@inproceedings{shao2023prompting,
  title={Prompting large language models with answer heuristics for knowledge-based visual question answering},
  author={Shao, Zhenwei and Yu, Zhou and Wang, Meng and Yu, Jun},
  booktitle={CVPR},
  year={2023}
}

@inproceedings{agrawal2019nocaps,
  title={Nocaps: Novel object captioning at scale},
  author={Agrawal, Harsh and Desai, Karan and Wang, Yufei and Chen, Xinlei and Jain, Rishabh and Johnson, Mark and Batra, Dhruv and Parikh, Devi and Lee, Stefan and Anderson, Peter},
  booktitle={ICCV},
  year={2019}
}

@inproceedings{hudson2019gqa,
  title={Gqa: A new dataset for real-world visual reasoning and compositional question answering},
  author={Hudson, Drew A and Manning, Christopher D},
  booktitle={CVPR},
  year={2019}
}

@inproceedings{lin2014microsoft,
  title={Microsoft coco: Common objects in context},
  author={Lin, Tsung-Yi and Maire, Michael and Belongie, Serge and Hays, James and Perona, Pietro and Ramanan, Deva and Doll{\'a}r, Piotr and Zitnick, C Lawrence},
  booktitle={ECCV},
  year={2014},
}

@misc{vicuna2023,
    title = {Vicuna: An Open-Source Chatbot Impressing GPT-4 with 90\%* ChatGPT Quality},
    url = {https://lmsys.org/blog/2023-03-30-vicuna/},
    author = {Chiang, Wei-Lin and Li, Zhuohan and Lin, Zi and Sheng, Ying and Wu, Zhanghao and Zhang, Hao and Zheng, Lianmin and Zhuang, Siyuan and Zhuang, Yonghao and Gonzalez, Joseph E. and Stoica, Ion and Xing, Eric P.},
    month = {March},
    year = {2023}
}

@inproceedings{lin2022revive,
  title={Revive: Regional visual representation matters in knowledge-based visual question answering},
  author={Lin, Yuanze and Xie, Yujia and Chen, Dongdong and Xu, Yichong and Zhu, Chenguang and Yuan, Lu},
  booktitle={NeurIPS},
  year={2022}
}

@inproceedings{lin2024preflmr,
  title={PreFLMR: Scaling Up Fine-Grained Late-Interaction Multi-modal Retrievers},
  author={Lin, Weizhe and Mei, Jingbiao and Chen, Jinghong and Byrne, Bill},
  booktitle={ACL},
  year={2024}
}

@inproceedings{lin2022retrieval,
  title={Retrieval Augmented Visual Question Answering with Outside Knowledge},
  author={Lin, Weizhe and Byrne, Bill},
  booktitle={EMNLP},
  year={2022}
}

@inproceedings{brown2020language,
  title={Language models are few-shot learners},
  author={Tom Brown and Benjamin Mann and Nick Ryder and Melanie Subbiah and Jared D Kaplan and Prafulla Dhariwal and Arvind Neelakantan and Pranav Shyam and Girish Sastry and Amanda Askell et al},
  booktitle={NeurIPS},
  year={2020}
}

@article{kwiatkowski2019natural,
  title={Natural questions: a benchmark for question answering research},
  author={Kwiatkowski, Tom and Palomaki, Jennimaria and Redfield, Olivia and Collins, Michael and Parikh, Ankur and Alberti, Chris and Epstein, Danielle and Polosukhin, Illia and Devlin, Jacob and Lee, Kenton and others},
  journal={Transactions of the Association for Computational Linguistics},
  year={2019},
}

@article{ram2023context,
  title={In-context retrieval-augmented language models},
  author={Ram, Ori and Levine, Yoav and Dalmedigos, Itay and Muhlgay, Dor and Shashua, Amnon and Leyton-Brown, Kevin and Shoham, Yoav},
  journal={TACL},
  year={2023},
}

@inproceedings{caffagni2024wiki,
  title={Wiki-LLaVA: Hierarchical Retrieval-Augmented Generation for Multimodal LLMs},
  author={Caffagni, Davide and Cocchi, Federico and Moratelli, Nicholas and Sarto, Sara and Cornia, Marcella and Baraldi, Lorenzo and Cucchiara, Rita},
  booktitle={CVPR Workshop},
  year={2024}
}

@inproceedings{yoranmaking,
  title={Making Retrieval-Augmented Language Models Robust to Irrelevant Context},
  author={Yoran, Ori and Wolfson, Tomer and Ram, Ori and Berant, Jonathan},
  booktitle={The Twelfth International Conference on Learning Representations},
  year={2024}
}

@article{chen2024expanding,
  title={Expanding performance boundaries of open-source multimodal models with model, data, and test-time scaling},
  author={Chen, Zhe and Wang, Weiyun and Cao, Yue and Liu, Yangzhou and Gao, Zhangwei and Cui, Erfei and Zhu, Jinguo and Ye, Shenglong and Tian, Hao and Liu, Zhaoyang and others},
  journal={arXiv preprint arXiv:2412.05271},
  year={2024}
}

@article{yang2024qwen2,
  title={Qwen2. 5 Technical Report},
  author={Yang, An and Yang, Baosong and Zhang, Beichen and Hui, Binyuan and Zheng, Bo and Yu, Bowen and Li, Chengyuan and Liu, Dayiheng and Huang, Fei and Wei, Haoran and others},
  journal={arXiv preprint arXiv:2412.15115},
  year={2024}
}

@inproceedings{ko2025bidirectional,
  title={Bidirectional likelihood estimation with multi-modal large language models for text-video retrieval},
  author={Ko, Dohwan and Lee, Ji Soo and Choi, Minhyuk and Meng, Zihang and Kim, Hyunwoo J},
  booktitle={ICCV},
  year={2025}
}

@inproceedings{lee2025vidchain,
  title={VidChain: Chain-of-Tasks with Metric-based Direct Preference Optimization for Dense Video Captioning},
  author={Lee, Ji Soo and Kim, Jongha and Na, Jeehye and Park, Jinyoung and Kim, Hyunwoo J},
  booktitle={AAAI},
  year={2025}
}

@inproceedings{kim2026tabflash,
  title={TabFlash: Efficient Table Understanding with Progressive Question Conditioning and Token Focusing},
  author={Kim, Jongha and Bae, Minseong and Lee, Sanghyeok and Yoon, Jinsung and Kim, Hyunwoo J},
  booktitle={AAAI},
  year={2026}
}

@inproceedings{park2024generative,
  title={Generative Subgraph Retrieval for Knowledge Graph--Grounded Dialog Generation},
  author={Park, Jinyoung and Joo, Minseok and Kim, Joo-Kyung and Kim, Hyunwoo},
  booktitle={EMNLP},
  year={2024}
}

@inproceedings{ko2023large,
  title={Large language models are temporal and causal reasoners for video question answering},
  author={Ko, Dohwan and Lee, Ji and Kang, Woo-Young and Roh, Byungseok and Kim, Hyunwoo},
  booktitle={EMNLP},
  year={2023}
}
